\documentclass[10pt,twocolumn,letterpaper]{article}

\usepackage[preprint]{cvpr}      
\usepackage{booktabs}       
\usepackage{amsfonts}       
\usepackage{nicefrac}       
\usepackage{xcolor}         
\usepackage{amsmath} 
\usepackage{multirow}
\usepackage{colortbl}
\usepackage{graphicx}
\usepackage{float}
\usepackage{colortbl}
\usepackage[accsupp]{axessibility}  

\newcommand{\method}{\textit{3DThinker}\xspace}

\definecolor{cvprblue}{rgb}{0.21,0.49,0.74}
\usepackage[pagebackref,breaklinks,colorlinks,allcolors=cvprblue]{hyperref}

\def\paperID{3615} 
\def\confName{CVPR}
\def\confYear{2026}

\title{Think with 3D: Geometric Imagination Grounded \\ Spatial Reasoning from Limited Views}

\author{
   Zhangquan Chen$^{1}$\thanks{\*The work was conducted during the internship of Zhangquan Chen (czq23@mails.tsinghua.edu.cn) at Meituan.} $\quad{}$  Manyuan Zhang$^{2}$\footnotemark[2] $\quad{}$  Xinlei Yu$^{3}$ $\quad{}$ Xufang Luo$^{4}$ $\quad{}$  Mingze Sun$^{1}$ $\quad{}$  Zihao Pan$^{2}$ \\
   Xiang An$^{5}$ $\quad{}$ Yan Feng$^{2}$ $\quad{}$ Peng Pei$^{2}$ $\quad{}$ Xunliang Cai$^{2}$ $\quad{}$ Ruqi Huang$^{1}$\thanks{Corresponding author: ruqihuang@sz.tsinghua.edu.cn, zhangmanyuan@link.cuhk.edu.hk}\\
   1. Tsinghua Shenzhen International Graduate School, Tsinghua University \\
   2. Meituan $\quad{}$
   3. National University of Singapore $\quad{}$
   4. Beihang University $\quad{}$
   5. LMMs-Lab
}

\begin{document}
\maketitle
\begin{abstract} 
Though recent advances in vision–language models (VLMs) have achieved remarkable progress across a wide range of multimodal tasks, understanding 3D spatial relationships from limited views remains a significant challenge. Previous reasoning methods typically rely on pure text (e.g., topological cognitive maps) or on 2D visual cues. However, their limited representational capacity hinders performance in specific tasks that require 3D spatial imagination. To address this limitation, we propose \method, a framework that can effectively exploit the rich geometric information embedded within images while reasoning, like humans do. Our framework is the first to enable 3D mentaling during reasoning without any 3D prior input, and it does not rely on explicitly labeled 3D data for training. Specifically, our training consists of two stages. First, we perform supervised training to align the 3D latent generated by VLM while reasoning with that of a 3D foundation model (e.g., VGGT). Then, we optimize the entire reasoning trajectory solely based on outcome signals, thereby refining the underlying 3D mentaling.
Extensive experiments across multiple benchmarks show that \method consistently outperforms strong baselines and offers a new perspective toward unifying 3D representations into multimodal reasoning. Our code is available at \url{https://github.com/zhangquanchen/3DThinker}.

\end{abstract}
\section{Introduction}
\label{sec:intro}
Spatial understanding is a critical capability for machines to interact with the real 3D world (e.g., embodied AI, autonomous driving)~\cite{wang2024embodiedscan,xia2018gibson, zeng2025FSDrive, yan2025drivingsphere}. 
These systems typically rely on ego-centric, multi-view observations, typically provided by multiple cameras simultaneously capturing limited views of their surroundings. These views are not interchangeable or purely visual; they inherently carry spatial semantics tied to the machine’s frame of reference~\cite{ego3d}. Consequently, imagining the full scene and performing reasoning based on a few limited views presents an essential problem for spatial intelligence~\cite{mirage}. 
Although recent VLMs are pretrained on large-scale image–text corpora, their performance on such spatial reasoning tasks remains notably limited~\cite{chen2024spatialvlmintro, cheng2024spatialrgptintro, liao2024spatialreasoningintro, yan2025olidm}. The core bottleneck lies in their \emph{inability to extract 3D geometry embedded within images} and their \emph{restricted capacity for spatial imagination}.

\begin{figure}[!t]
    \centering
    \includegraphics[width=0.45\textwidth]{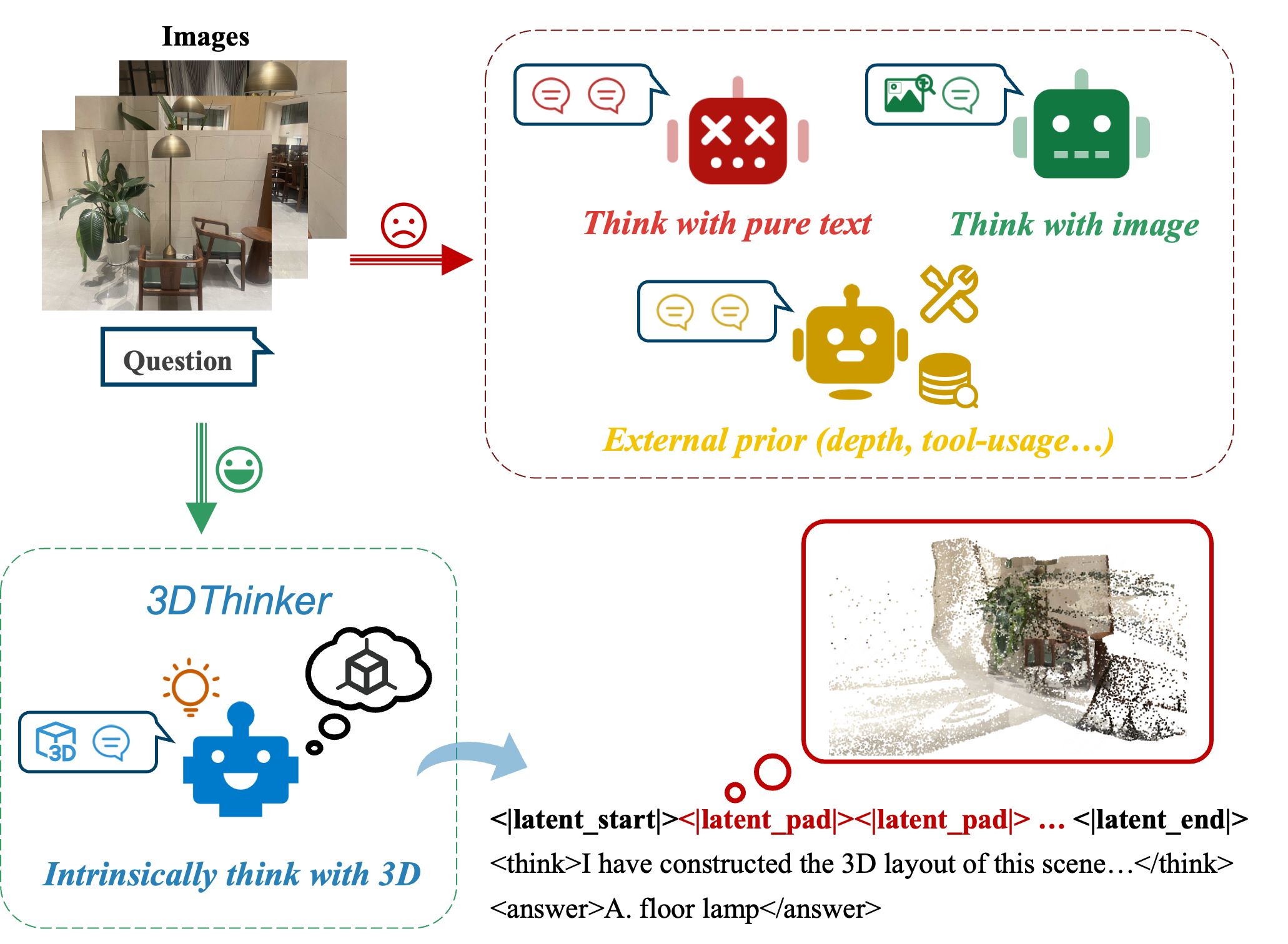}
    \vspace{-0.5em}
    \caption{Illustration of our \method. Existing methods typically perform reasoning based solely on pure text or 2D visual cues, without fully exploiting the rich spatial and geometric information inherent in images. Other methods attempt to enhance the input by introducing auxiliary modalities (e.g., depth maps or coordinates), yet these often depend on additional annotations or external tools. In contrast, our framework enables VLMs to intrinsically form 3D mental representations during reasoning, thereby improving their spatial understanding.}\label{fig:teaser}
    \vspace{-1.8em}
\end{figure}
Recent advances have attempted to enhance the spatial reasoning capabilities of VLMs~\cite{ouyang2025spacer,rela:liu2025spatialcot,rela_new:mideye,li2025spatialladder,ego3d,liu2025ssr,chen2025sifthinker}. As illustrated in Fig.~\ref{fig:teaser}, existing methods can be broadly divided into two categories. The first category performs reasoning with pure text~\cite{chen2024spatialvlm,liu2025ssr,cheng2024spatialrgpt,ouyang2025spacer,rela:liu2025spatialcot} or 2D visual cues~\cite{mmspatial, chen2025sifthinker, xiao2024towards}, whose representational capacity for complex spatial layouts is inherently limited. To mitigate this limitation, methods such as MindCube~\cite{mindcube} train models to generate cognitive maps of 3D layouts; however, they rely on bird’s-eye-view (BEV) annotations to construct these maps. 
Ego3D~\cite{ego3d} further employs external models—GroundingDINO~\cite{groundingdino} for referring expression comprehension (REC) and DepthAnythingv2~\cite{yang2024depth} for depth estimation, to automatically generate cognitive maps
Yet, constrained by the performance of these models, such methods often fail on low-resolution or uncurated images.
The second category incorporates auxiliary modalities as additional inputs (e.g., point clouds, camera parameters)~\cite{chen2025vidbotprior,li2025pointvlaprior}. However, these settings restrict the model’s applicability in real-world scenarios where only monocular images are available. Moreover, several recent methods invoke external encoders or tool-usage to obtain prior information (e.g., encoded 3D tokens~\cite{vlm3r}, depth maps~\cite{liu2025ssr,chen2025sifthinker,cai2024spatialbot}). Importantly, these techniques do not constitute an intrinsic capability of the model and introduce additional inference overhead.

These challenges motivate the need for a new method that: G1) \textbf{3D-imaginable}: can directly learn 3D geometry from limited 2D images; G2) \textbf{Annotation-free}: does not rely on densely annotated data; and G3) \textbf{Intrinsic}: requires no external priors or auxiliary models during inference.

The most relevant mental model, Mirage~\cite{mirage}, leverages ground-truth image embeddings for supervised training, facilitating the continuation of a multimodal trajectory without the need for pixel-level image generation. 
However, the training of~\cite{mirage} is heavily reliant on ground-truth image supervision and remains constrained to the "thinking with image" paradigm, which prevents its effectiveness on (G1) and (G2). 
Nevertheless, it provides a valuable inspiration, prompting us to \emph{introduce a new novel framework, \method, which enables thinking with 3D mentaling}. Unlike prior works that depend on external priors or complex training data construction, our method intrinsically integrates 3D representations into the VLMs, enabling unified reasoning and 3D latent generation within the model. For (G1), our framework enables the model to generate geometric representations from images during the reasoning process. Regarding (G2), we directly project the 3D latent to align with a 3D foundation model, thereby circumventing the need for raw 3D data construction. Consequently, our model can inherently "think with 3D" without relying on any prior or auxiliary geometry encoder, corresponding to (G3).
Simultaneously, since our method allows for the recovery of 3D representations(e.g., point clouds) from 3D latents via the projector, it \emph{significantly enhances the interpretability of the large reasoning model}.

Specifically, we first construct a batch of Chain-of-Thought (CoT) data that incorporates 3D special tokens. Our training framework then proceeds in two main stages. In the first stage, we perform supervised learning, where features from the 3D foundation model (e.g., VGGT~\cite{wang2025vggt}) are distilled into the native reasoning process of the VLM. To enable the model to think with a 3D mentaling while maintaining textual coherence, we employ both 3D latent alignment loss and the cross-entropy loss. In the second stage, we employ reinforcement learning, optimizing the tokens across the entire sampling trajectory based solely on outcome-driven signals, while preserving the alignment of the 3D latent. That is, we refine 3D mentaling within the trajectory using only outcome as the optimization signal.

Our contributions can be summarized as follows.

\begin{figure*}[]
    \centering
    \includegraphics[width=\textwidth]{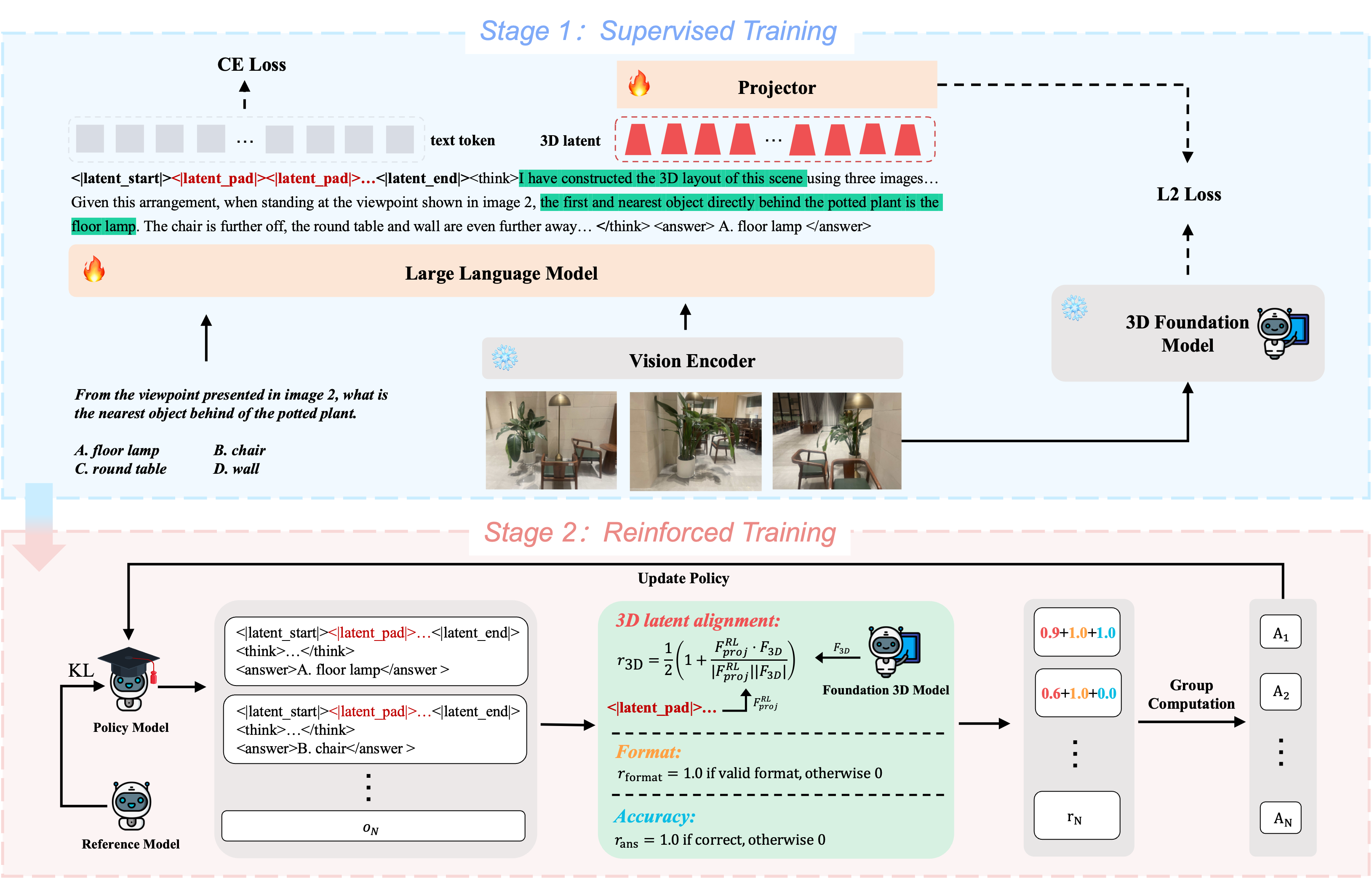}
    \vspace{-1.5em}
    \caption{The schematic illustration of our \method, a framework that enables thinking with 3D mentaling. (1) Stage 1: \method is first trained under supervision using our constructed CoT data (see Sec.~\ref{sec:data}), aligning the generated 3D latents with the feature space of 3D foundation model. This alignment allows the model to leverage suitable 3D spatial mentaling while reasoning. (2) Stage 2: After supervised training, we further optimize the entire trajectory using only outcome signals, while maintaining the alignment of the 3D latents.}\label{fig:pipeline}
    \vspace{-1.5em}
\end{figure*}

\begin{itemize}
\item We are the first to introduce the "think with 3D mentaling" framework, which operates without dependence on densely labeled training data (e.g., cognitive maps).

\item We propose a two-stage training framework (shown in Fig.~\ref{fig:pipeline}), progressing from feature alignment to learning intrinsic geometry awareness from outcome-based signals, thus enabling 3D mentaling without any external prior.

\item \method overcomes the lack of interpretability in latent reasoning. Specifically, \method enables the recovery of 3D representations from the latent space via a projector during the reasoning process.

\item Extensive experiments across multiple benchmarks demonstrate that \method consistently outperforms strong baselines. Furthermore, our results indicate that the effectiveness of \method generalizes well across different base VLMs, highlighting its broad applicability.
\end{itemize}
\section{Related Work}
\label{sec:related}
\subsection{Multimodal Reasoning}
Large language models (LLMs) have experienced rapid development, and demonstrated strong performance across a wide range of tasks~\cite{chen2024three, zhou2025foodsky, zeng2025bridgingeditinggapllms, sec, yu2025forgetme, xiang2025promptsculptor, chu2025safekv, yao2024swift}. Building on these advances, recent works have highlighted that in-context learning, including intermediate rationales, can significantly enhance the performance of LLMs~\cite{sarkar2025reasoning, cotchain,cottheory, wang2025reasoning, lin2025planbudgeteffectiveefficient, zhou2025valuing, yu2025physics}. Current reasoning methods can be categorized into three types: pure-text, visual, and latent reasoning. 

\textbf{Pure-text reasoning:} ~\cite{rela_new:r1v, rela_new:llamavo1, rela_new:alignanything, rela_new:openrlhf, shen2025vlmr1, univg-r1} elicit textual step-by-step reasoning inspired by \cite{rela:deepseekr1}. They typically rely on textual descriptions, which can limit the reasoning capabilities when dealing with visual evidence that cannot be adequately described using pure textual language. \textbf{Visual reasoning:} to solve the problem mentioned above, some methods integrate visual evidence directly into the reasoning trajectory, whether in multi-hop or continuous modes. Some intrinsic multi-hop methods~\cite{rela:liu2025spatialcot, shao2024visual, chen2025visrl, rela:wang2024segllm}, first generate detailed visual cues within the model itself (e.g., bounding boxes, coordinates, or masks), and then the further reasoning is conducted based on these cues. Other extrinsic tool-usage methods~\cite{openai2025o3o4mini, su2025openthinkimg, wu2025vtool, zheng2025deepeyes}, enhance the "think with image" capability by dynamically invoking external image tools. On the other hand, continuous methods like GRIT~\cite{fan2025grit} and SIFThinker~\cite{chen2025sifthinker} generate continuous visual reasoning to enable iterative corrections during the single-step reasoning process. \textbf{Latent reasoning:} some studies have shown that incorporating intermediate hidden representations into LLMs can effectively enhance model capabilities~\cite{biran2024hopping, yang2024large, deng2024explicit, sun2025audioenhanced}. ~\cite{hao2024training} replaces CoT tokens with continuous latent embeddings, allowing unconstrained reasoning in the latent space to tackle complex tasks. More recently, Mirage~\cite{mirage} and LVR~\cite{lvr} utilize special visual tokens alongside ordinary text during reasoning. They explore visual information within the model by implicitly supervising the generation of image latent, thereby enabling reasoning with 2D visual latent. 

While prior works primarily focus on enhancing reasoning ability in textual or 2D spaces, our method takes a different perspective: \emph{we treat latent tokens as a bridge for the model to think with 3D at a mental-level, aligning more closely with human cognition.}

\subsection{Spatial Understanding}
Spatial understanding encompasses skills such as 3D imagination and spatial cognition, which are essential for perceiving and manipulating spatial relationships in both 2D and 3D environments~\cite{xu2025defining, zha2025enable, cai2025has, ni2025recondreamer, wang2025embodiedreamer, zeng2025janusvln, ni2025wonderturbo, zhang2024yoloppa, ding2025neural, yao2023depthssc}. Recently, many efforts have been dedicated to evaluating the spatial understanding ability of VLMs~\cite{ego3d, mindcube, ray2024sat, zhang2025open3dvqa, ma20243dsrbench, zhang2024sphere, lee2025perspective}. Additionally, several methods have been proposed to enhance spatial understanding. 
For example, ~\cite{liu2025ssr, cai2024spatialbot, mmspatial,ouyang2025spacer, qi2025gpt4scene} equip LLM with additional multiview, depth or point cloud inputs, essentially serving as input enhancement. Furthermore, 3DRS~\cite{3drs} introduces a teacher model for 3D supervision to achieve explicit spatial representation alignment; however, this method requires input that includes the 3D coordinates corresponding to each pixel. Moreover, VLM-3R~\cite{vlm3r} employs implicit 3D tokens from a pre-trained model (e.g., CUT3R~\cite{cut3r}) to achieve spatial awareness by incorporating prior information, necessitating inference with an extensively 3D foundation model. Recently, methods like MindCube~\cite{mindcube} and Ego3D-VLM~\cite{ego3d} have facilitated spatial understanding by constructing textual cognitive maps. 

Despite these advancements, existing methods often rely on input enhancement or constructed cognitive maps, necessitating complex data collection and annotation. However, \emph{\method enables 3D mentaling directly from multi views by learning 3D latent distilled from 3D foundation models, thereby facilitating spatial reasoning without relying on densely annotated data.}
\section{Methodology}
\label{sec:method}
Human cognition is inherently rooted in the comprehension of 3D environments. Inspired from the cognitive mechanism of mental imagery, we propose \method, \emph{a framework that enables VLMs to imagine 3D scenes during reasoning processes}. In contrast to existing methods that reason with pure text or 2D visual cues, our framework integrates 3D representations into the interleaved multimodal trajectories. Specifically, \method generates compact latent embeddings that serve as 3D tokens, closely emulating the mental 3D scenes that humans intuitively imagine in spatial reasoning. As illustrated in Fig.~\ref{fig:pipeline}, \method first aligns the VLM-generated 3D latent with the 3D foundation model, followed by reinforced training to optimize the trajectory. In this section, we will explain how we achieve this from three aspects: data generation, supervised training (stage 1), and reinforcement training (stage 2).

\subsection{Data Generation}
\label{sec:data}
Due to the fact that VLMs naturally only generate textual tokens, they require additional supervised training to learn how to \emph{produce interleaved reasoning patterns that incorporate 3D information}. Therefore, we synthesize specific training corpora based on the raw question-answer data accompanied by multi-view images.
Given the image set $\mathcal{I} = \{I_1, I_2, \ldots, I_n\}$, a question $Q$, and the ground truth response $R$, we employ a high-level model (i.e., GPT-4.1) $M$ to complete the reasoning chain. Specifically, we prompt the model $M$ to generate step-by-step reasoning that contains placeholders (3D special tokens), where these tokens represent imagined 3D scenes in the mind. Denote the response $o$ as:
\begin{equation}
o = M(Q, \mathcal{I}, R).
\end{equation}

Here, $o$ represents the step-by-step reasoning process with embedded 3D placeholders, whose last layer hidden states are required to be consistent with features extracted from the 3D foundation model during supervised training. By prompting the large-scale reasoning VLM with various inputs, we are able to collect a training dataset $\mathcal{D} = \left\{ \left( Q^{(i)}, \mathcal{I}^{(i)}, R^{(i)}, o^{(i)} \right) \right\}$, where each $o^{(i)}$ contains interleaved text and 3D placeholders.

\subsection{Supervision for 3D Grounded Reasoning}
\label{sec:sft}
To teach the model reasoning with 3D, a naive solution is to explicitly align its outputs with 3D representations (e.g., point cloud). However, this often depends on labor-intensive data annotation and requires the model to have explicit 3D generation capabilities, which can be quite challenging. Instead, we introduce the 3D foundation model (i.e., VGGT~\cite{wang2025vggt}) during training, and distill its features to the 3D special token generated within the VLM reasoning process, thereby facilitating effective \emph{3D-aware reasoning without the need for exhaustive manual labeling}.

Specifically, for each training example $\left( Q, \mathcal{I}, R, o \right) \in \mathcal{D}$, the reasoning trajectory $o$ can be decomposed into three sequential components through concatenation operations:
\begin{equation}
    o = o_{\text{pre}} \oplus t_{\text{3D}} \oplus o_{\text{post}},
\end{equation}
where $t_{\text{3D}} = \{t_1, \ldots, t_k\}$ represents the token sequence of human-like 3D mental imagery. The salient vectors $F_{\text{latent}} = \{h_1, \ldots, h_k\}$, which operationalize the 3D cognitive tokens $t_{\text{3D}}$, are extracted from the last layer hidden states of VLM $f_\theta(\cdot)$ with parameter $\theta$. These salient vectors are recursively generated conditioned on the preceding context:
\begin{equation}
    h_i = 
    \begin{cases}
        f^{\mathrm{hidden},L}_{\theta}(Q, \mathcal{I}, o_{\text{pre}}), & i = 1, \\
        f^{\mathrm{hidden},L}_{\theta}(Q, \mathcal{I}, o_{\text{pre}}, t_{1:i-1}), & i \geq 2.
    \end{cases}
\end{equation}

\begin{figure}[]
    \centering
    \includegraphics[width=0.5\textwidth]{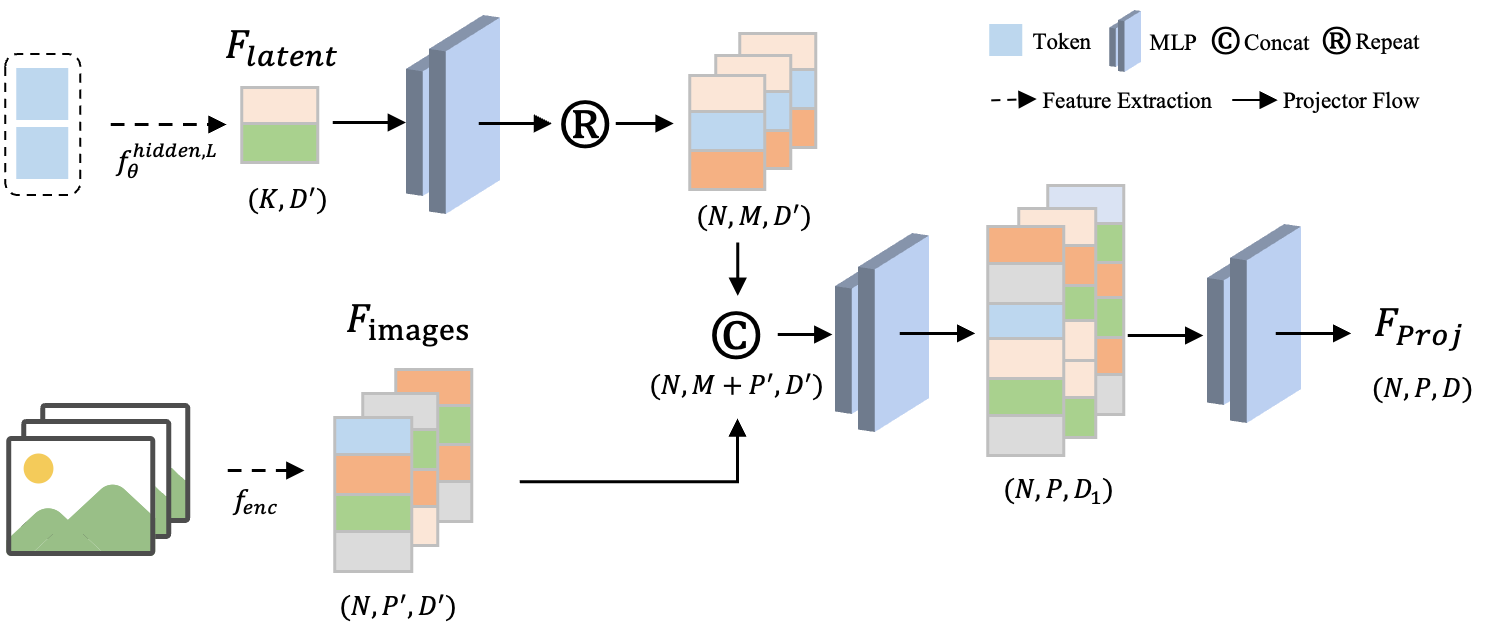}
    \vspace{-1.2em}
    \caption{Illustration of our projector, which transforms VLM-generated 3D latent into the feature space of VGGT.}\label{fig:projector}
    \vspace{-1.5em}
\end{figure}

Concurrently, we can obtain patch-level visual features $F_{\text{images}} = f_{\text{enc}}(\mathcal{I})$ from the image encoder, and acquire the geometry features $F_{\text{3D}} = f_{\text{vggt}}(I)$ through the last layer of VGGT aggregator. 
To ensure dimensional consistency between the generated 3D latent features and the predicted geometry features, we employ the projector as illustrated in Fig.~\ref{fig:projector} to transform $F_{\text{latent}}$ into a compatible feature space:
\begin{equation}
\label{eq:proj}
    F_{\text{proj}} = Projector(F_{\text{latent}}, F_{\text{images}}).
\end{equation}

Our objective is to achieve optimal alignment between the projected 3D features derived from the VLM and the corresponding VGGT features. To this end, we formulate the 3D alignment as the Frobenius loss:
\begin{equation}
    \mathcal{L}_{3D} = \| F_{\text{proj}} - F_{\text{3D}} \|_F^2.
\end{equation}

On the other hand, to ensure textual coherence while introducing 3D tokens, we employ cross-entropy loss to optimize the prediction of surrounding textual tokens. Specifically, the prediction of $i-th$ textual tokens before is $t_{\text{3D}}$ conditioned on both the preceding response tokens and the original input sequence.

\begin{equation}
    \mathcal{L}_{\text{text}}^{\text{pre}}=\sum_{i=1}^{\left|o_{\text{pre}}\right|} \ell_{\mathrm{CE}}\left(o_{\text {pre},i}, f_\theta\left(Q,\mathcal{I}, o_{\text {pre},<i}\right)\right).
\end{equation}
In contrast, textual tokens positioned after $t_{\text{3D}}$ incorporates $k$ textual 3D special tokens.
\begin{equation}
    \mathcal{L}_{\text{text}}^{\text{post}}=\sum_{i=1}^{\left|o_{\text {post}}\right|} \ell_{\mathrm{CE}}\left(o_{\text {post},i}, f_\theta\left(Q,\mathcal{I}, o_{\text {pre}},t_{\text{3D}}, o_{\text {post},<i}\right) .\right.
\end{equation}
Finally, the textual loss is formulated as follows:
\begin{equation}
    \mathcal{L}_{\text{text}}=\mathcal{L}_{\text{text}}^{\text{pre}} + \mathcal{L}_{\text{text}}^{\text{post}}.
\end{equation}

The overall training objective incorporates both 3D alignment and textual losses, thereby enabling the model to seamlessly incorporate 3D imaging into its textual reasoning process. Here, $\lambda_{3D}$ and $\lambda_{text}$ serve as hyperparameters that balances coefficients.
\begin{equation}
\label{eq:sft}
    \mathcal{L}_{\text{total}}=\lambda_{3D}\mathcal{L}_{3D} + \lambda_{text}\mathcal{L}_{text}.
\end{equation}

\subsection{Reinforced Spatial Mentaling}
\label{sec:rl}
At the supervised training stage, our primary objective is to enable the model to perform textual reasoning while simultaneously generating formatted 3D tokens. Additionally, we pre-train the projector to achieve effective alignment of 3D latents. During the reinforced training stage, we expect to use \emph{only outcome signals} to optimize the sampling trajectories and refine the imagined mental 3D representations as well. Specifically, we employ outcome-based group-relative policy optimisation (GRPO)~\cite{shao2024deepseekmath}, while VGGT features are utilized to further optimize the 3D visual token generated by the model. Notably, the projector remains frozen in this stage. We formalize the RL framework as follows.

For each question-images pair $(Q, \mathcal{I})$, the reinforcement learning (RL) framework generates a set of candidate completions $\{o_1, \ldots, o_N\}$ from the current policy $\pi_{\theta_\text{old}}$, and subsequently updates the policy to $\pi_\theta$ by maximizing the following objective:
\begin{equation}
    \begin{aligned}
    \mathcal{J}(\theta) = &
    \frac{1}{N} \sum_{i=1}^N \frac{1}{|o_i|} \sum_{t=1}^{|o_i|} \Big\{ \min \Big[
    \text{clip}(r_{i,t}, 1 - \epsilon, 1 + \epsilon)\hat{A}_{i,t},\; 
    \\&r_{i,t} \hat{A}_{i,t}\Big] - \beta\, \mathbb{D}_\text{KL}[\pi_\theta \,||\, \pi_\text{ref}]\Big\},
    \end{aligned}
\end{equation}
where $r_{i,t} = \frac{\pi_\theta(o_{i,t} \mid q, o_{i,<t})}{\pi_{\theta_\text{old}}(o_{i,t} \mid q, o_{i,<t})}$ denotes the likelihood ratio between the updated and old policies at step $t$. $\epsilon, \beta$ are hyperparameters, and $\mathbb{D}_\text{KL}[\pi_\theta \,||\, \pi_\text{ref}]$ represents the KL divergence~\cite{schulman2020approximating} between the current policy model and the fixed reference model. 

The group-normalized advantage, denoted as $\hat{A}_{i,t}$, is calculated by the task-specific reward $r_{i,t}$.
\begin{equation}
    \hat{A}_{i,t} = \frac{r_{i,t} - \text{mean}\{r_{1,t}, \ldots, r_{N,t}\}}{\text{std}\{r_{i,t}, \ldots, r_{N,t}\} + \delta}.
\end{equation}

Next, we will introduce several specifically designed rewards to achieve reinforced spatial mentaling.

\paragraph{Reward for 3D visual token.} 
After the supervised training, the model has begun to exhibit the ability of 3D mentaling during the thinking process. To further optimize the 3D visual token in the reasoning process, we can extract the last layer hidden state of 3D special token $t_{3D}$ (i.e.,  {\small \texttt{<|latent\_start|><|latent\_pad|> ... <|latent\_end|>}}) in each trajectory $o$ and perform optimization. Specifically, the projected features $F_{proj}^{RL}$ are computed based on Eq.~\ref{eq:proj} at step $t$, with VGGT features $F_{3D}$ serving as constraints during the RL stage. That is, the cosine similarity between the VGGT features and the projected features is calculated to serve as the reward $r_{\text{3D}}$.
\begin{equation}
    r_{\text{3D}} = \frac{1}{2}(1 + \frac{{F_{proj}^{RL} \cdot F_{3D}}}{{\|F_{proj}^{RL}\| \|F_{3D}}\|}),
\end{equation}

\paragraph{Reward for outcome-based optimization.} 
\label{sec:reward}
We expect to optimize the entire trajectory using only the outcome-based signals, without relying on explicit annotations of intermediate processes. Thus, we design corresponding rewards for both format ($r_{\text{format}}$) and final answer ($r_{\text{ans}}$). \textbf{(1) Format reward:} the model's output should adhere to the format: {\small \texttt{...<|latent\_start|><|latent\_pad|>... <|latent\_end|>...<think>...</think><answer> ... </answer>}}. A reward of 1.0 is assigned to responses that strictly comply with this format. \textbf{(2) Answer reward:} we also provide the 0/1 binary reward by comparing the generated answer with the ground truth option. This outcome-based reward is evenly distributed across each token in the trajectory, including 3D visual tokens.

So, the task reward $r_{i,t}$ is a composite signal comprising the sum of three components: $r_{\text{3D}}, r_{\text{format}}$ and $r_{\text{ans}}$.
\section{Experiments}
\label{sec:exp}
\begin{table*}[t!]
    \caption{Accuracy comparison of generalist VLMs and our method (\method) on MindCube-Tiny and Ego3D-Bench, with our training conducted on stage 1 (S1) and on both stage 1 and stage 2 (S1 + S2). The best results achieved based on different VLMs are \textbf{bolded}. The overall/average results of each model are highlighted in \textcolor{blue}{blue}, with the best results among all models highlighted in \textbf{\textcolor{red}{red}}.}\label{table:mllm}
    \vspace{-0.8em}
    \centering
    \footnotesize
    \setlength{\tabcolsep}{1.4mm}
    \begin{tabular}{lcccccccccccccc}
    \hline
                                      & \multicolumn{4}{c}{\textbf{MindCube-Tiny}}                               & \textbf{} & \multicolumn{9}{c}{\textbf{Ego3D-Bench}}                                                                                                                                                                                                                                                                                                                                                                                                                                                          \\ \cline{2-5} \cline{7-15} 
    \multirow{-2}{*}{\textbf{Method}} & \textbf{Rotation} & \textbf{Among} & \textbf{Around} & {\color[HTML]{0606FF} \textbf{Overall ↑}}     & \textbf{} & \textbf{\begin{tabular}[c]{@{}c@{}}Ego \\ Dist.\end{tabular}} & \textbf{\begin{tabular}[c]{@{}c@{}}Obj.\\ Dist.\end{tabular}} & \textbf{Loc.} & \textbf{\begin{tabular}[c]{@{}c@{}}Ego \\ Mot.\end{tabular}} & \textbf{\begin{tabular}[c]{@{}c@{}}Obj. \\ Mot.\end{tabular}} & \textbf{\begin{tabular}[c]{@{}c@{}}Travel\\ Time\end{tabular}} & \textbf{\begin{tabular}[c]{@{}c@{}}Ego \\ Rel.\end{tabular}} & \textbf{\begin{tabular}[c]{@{}c@{}}Obj. \\ Rel.\end{tabular}} & {\color[HTML]{0606FF} \textbf{Avg.↑}} \\ \hline
    \multicolumn{15}{c}{{\color[HTML]{000000} \textit{Closed-source Models}}}                                                                                                                                                                                                                                                                                                                                                                                                                                                                                                                                                                                          \\
gpt-4o-2024-11-20 & 37.0          & \textbf{44.8} & 56.4          & {\color[HTML]{0606FF} 46.1}          &  & 33.2          & 26.5          & 28.1          & 78.1          & 56.7          & 36.0          & 60.5          & 66.0          & {\color[HTML]{0606FF} 48.1}          \\
gpt-4.1           & 45.5          & 44.2          & 47.2          & {\color[HTML]{0606FF} 45.1}          &  & 51.7          & 36.2          & 41.8          & 82.7          & 62.6          & 44.3          & 65.7          & 70.2          & {\color[HTML]{0606FF} 56.9}          \\
glm-4.5v          & 28.0          & 43.0          & 33.2          & {\color[HTML]{0606FF} 37.8}          &  & 49.9          & 39.6          & 48.4          & 88.8          & 73.4          & 40.4          & 57.1          & 81.9          & {\color[HTML]{0606FF} 59.9}          \\
gemini-2.5-pro    & 84.0          & 39.7          & 56.8          & {\color[HTML]{0606FF} 52.2}          &  & 58.5          & 50.2          & 61.4          & 92.9          & 75.5          & 43.5          & 72.8          & 78.6          & {\color[HTML]{0606FF} 66.7}          \\
claude-sonnet-4   & 49.5          & 42.2          & 12.8          & {\color[HTML]{0606FF} 36.6}          &  & 48.9          & 36.5          & 51.6          & 81.9          & 55.1          & 33.6          & 53.9          & 69.5          & {\color[HTML]{0606FF} 53.9}          \\
doubao-seed-1.6   & \textbf{87.0} & 35.8          & 38.0          & {\color[HTML]{0606FF} 46.1}          &  & 55.2          & 50.8          & 60.5          & 89.0          & 67.3          & 49.8          & 71.4          & 86.0          & {\color[HTML]{0606FF} 66.3}          \\
o3-2025-04-16     & 86.5          & 42.7          & \textbf{66.0} & {\color[HTML]{0606FF} \textbf{56.6}} &  & \textbf{71.3} & \textbf{59.3} & \textbf{65.6} & \textbf{93.4} & \textbf{80.1} & \textbf{53.5} & \textbf{77.7} & \textbf{83.1} & {\color[HTML]{0606FF} \textbf{73.0}}         \\ \hline
    \multicolumn{15}{c}{{\color[HTML]{000000} \textit{Qwen2.5-VL Family}~\cite{bai2025qwen2.5}}}                                                                                                                                                                                                                                                                                                                                                                                                                                                                                                                                                                                             \\
    Qwen2.5-VL-3B                     & 37.4              & 33.3           & 30.3            & {\color[HTML]{0606FF} 33.2}          &           & 21.5                                                          & 29.4                                                          & 28.8          & 50.3                                                         & 41.9                                                          & \textbf{30.9}                                                  & 54.1                                                         & 56.1                                                          & {\color[HTML]{0606FF} 39.1}           \\
    $\textit{\method-S1}_\textnormal{Qwen2.5-3B}$              & 44.0              & 64.8           & 72.4            & {\color[HTML]{0606FF} 62.7}          &           & 36.1                                                          & 39.4                                                          & 32.5          & \textbf{54.8}                                                & 46.2                                                          & 30.8                                                           & 64.0                                                         & 69.7                                                          & {\color[HTML]{0606FF} 46.7}           \\
    \rowcolor[HTML]{E7E6E6} 
    $\textit{\method-S1+S2}_\textnormal{Qwen2.5-3B}$           & \textbf{55.5}     & \textbf{81.8}  & \textbf{75.2}   & {\color[HTML]{0606FF} \textbf{75.2}} &           & \textbf{41.6}                                                 & \textbf{46.0}                                                 & \textbf{33.1} & 54.7                                                         & \textbf{53.3}                                                 & 30.8                                                           & \textbf{70.1}                                                & \textbf{76.9}                                                 & {\color[HTML]{0606FF} \textbf{50.8}}  \\
    Qwen2.5-VL-7B                     & 36.5              & 32.5           & 38.4            & {\color[HTML]{0606FF} 34.7}          &           & 32.7                                                          & 31.5                                                          & 30.5          & 45.9                                                         & 44.0                                                          & {\color[HTML]{000000} 34.5}                                    & 43.2                                                         & 66.5                                                          & {\color[HTML]{0606FF} 41.1}           \\
    $\textit{\method-S1}_\textnormal{Qwen2.5-7B}$                  & 43.5              & 66.3           & \textbf{76.4}   & {\color[HTML]{0606FF} 64.4}          &           & 47.9                                                          & 44.5                                                          & \textbf{36.5} & 51.9                                                         & 51.3                                                          & \textbf{39.1}                                                  & 59.1                                                         & 73.9                                                          & {\color[HTML]{0606FF} 50.5}           \\
    \rowcolor[HTML]{E7E6E6} 
    $\textit{\method-S1+S2}_\textnormal{Qwen2.5-7B}$           & \textbf{55.0}     & \textbf{83.0}  & 76.0            & {\color[HTML]{0606FF} \textbf{76.0}} &           & \textbf{54.0}                                                 & \textbf{52.3}                                                 & \textbf{36.5} & \textbf{52.7}                                                & \textbf{56.6}                                                 & 38.2                                                           & \textbf{66.0}                                                & \textbf{83.1}                                                 & {\color[HTML]{0606FF} \textbf{54.9}}  \\
    Qwen2.5-VL-32B                    & 39.5              & 34.5           & 43.6            & {\color[HTML]{0606FF} 37.6}          &           & 45.4                                                          & 40.7                                                          & 49.6          & 75.6                                                         & 74.1                                                          & 40.1                                                           & 54.0                                                         & 79.0                                                          & {\color[HTML]{0606FF} 57.3}           \\
    $\textit{\method-S1}_\textnormal{Qwen2.5-32B}$                  & 45.0              & 66.8           & \textbf{77.2}   & {\color[HTML]{0606FF} 65.1}          &           & 52.0                                                          & 51.9                                                          & \textbf{54.8} & 80.1                                                         & 79.4                                                          & \textbf{44.3}                                                  & 62.0                                                         & 83.1                                                          & {\color[HTML]{0606FF} 63.5}           \\
    \rowcolor[HTML]{E7E6E6} 
    $\textit{\method-S1+S2}_\textnormal{Qwen2.5-32B}$           & \textbf{56.5}     & \textbf{83.2}  & \textbf{77.2}   & {\color[HTML]{0606FF} \textbf{76.7}} &           & \textbf{62.2}                                                 & \textbf{61.9}                                                 & 54.5          & \textbf{80.2}                                                & \textbf{86.6}                                                 & 43.7                                                           & \textbf{69.9}                                                & \textbf{86.0}                                                 & {\color[HTML]{0606FF} \textbf{68.1}}  \\
    Qwen2.5-VL-72B                    & 40.0              & 42.5           & 44.4            & {\color[HTML]{0606FF} 42.5}          &           & 42.4                                                          & 38.6                                                          & 54.8          & 86.8                                                         & 68.9                                                          & 38.5                                                           & 53.3                                                         & 80.5                                                          & {\color[HTML]{0606FF} 58.0}           \\
    $\textit{\method-S1}_\textnormal{Qwen2.5-72B}$                  & 42.5              & 68.0           & 73.6            & {\color[HTML]{0606FF} 64.5}          &           & 49.9                                                          & 45.9                                                          & 57.8          & 85.6                                                         & 75.6                                                          & \textbf{43.9}                                                  & 58.0                                                         & 80.8                                                          & {\color[HTML]{0606FF} 62.2}           \\
    \rowcolor[HTML]{E7E6E6} 
    $\textit{\method-S1+S2}_\textnormal{Qwen2.5-72B}$           & \textbf{57.0}     & \textbf{83.7}  & \textbf{77.6}   & {\color[HTML]{0606FF} \textbf{77.1}} &           & \textbf{61.1}                                                 & \textbf{59.9}                                                 & \textbf{59.7} & \textbf{93.1}                                                & \textbf{84.9}                                                 & 43.7                                                           & \textbf{69.8}                                                & \textbf{87.8}                                                 & {\color[HTML]{0606FF} \textbf{70.0}}  \\ \hline
    \multicolumn{15}{c}{{\color[HTML]{000000} \textit{InternVL3 Family}~\cite{zhu2025internvl3}}}                                                                                                                                                                                                                                                                                                                                                                                                                                                                                                                                                                                            \\
    InternVL3-8B                      & 37.0              & 40.3           & 63.2            & {\color[HTML]{0606FF} 45.1}          &           & 25.8                                                          & 28.7                                                          & 29.8          & 54.1                                                         & 54.8                                                          & 36.1                                                           & 49.9                                                         & 65.2                                                          & {\color[HTML]{0606FF} 43.1}           \\
    $\textit{\method-S1}_\textnormal{InternVL3-8B}$                  & 43.0              & 66.8           & \textbf{79.2}   & {\color[HTML]{0606FF} 65.2}          &           & 43.8                                                          & 44.4                                                          & 32.9          & 60.6                                                         & 61.2                                                          & \textbf{46.9}                                                  & 64.1                                                         & 72.1                                                          & {\color[HTML]{0606FF} 53.3}           \\
    \rowcolor[HTML]{E7E6E6} 
    $\textit{\method-S1+S2}_\textnormal{InternVL3-8B}$           & \textbf{55.0}     & \textbf{82.5}  & \textbf{79.2}   & {\color[HTML]{0606FF} \textbf{76.5}} &           & \textbf{54.6}                                                 & \textbf{56.1}                                                 & \textbf{36.0} & \textbf{67.2}                                                & \textbf{69.4}                                                 & 46.7                                                           & \textbf{71.0}                                                & \textbf{81.9}                                                 & {\color[HTML]{0606FF} \textbf{60.4}}  \\
    InternVL3-14B                     & 36.0              & 48.0           & 55.6            & {\color[HTML]{0606FF} 47.5}          &           & 46.0                                                          & 35.6                                                          & 35.9          & 63.2                                                         & 65.9                                                          & 41.6                                                           & 55.5                                                         & 70.1                                                          & {\color[HTML]{0606FF} 51.7}           \\
    $\textit{\method-S1}_\textnormal{InternVL3-14B}$                  & 42.0              & 68.3           & 77.2            & {\color[HTML]{0606FF} 65.4}          &           & 56.2                                                          & 49.1                                                          & 37.3          & 70.0                                                         & 71.8                                                          & \textbf{51.1}                                                  & 68.0                                                         & 77.7                                                          & {\color[HTML]{0606FF} 60.2}           \\
    \rowcolor[HTML]{E7E6E6} 
    $\textit{\method-S1+S2}_\textnormal{InternVL3-14B}$           & \textbf{54.5}     & \textbf{84.3}  & \textbf{77.6}   & {\color[HTML]{0606FF} \textbf{77.0}} &           & \textbf{63.5}                                                 & \textbf{59.9}                                                 & \textbf{41.3} & \textbf{78.3}                                                & \textbf{80.2}                                                 & 50.0                                                           & \textbf{75.1}                                                & \textbf{84.0}                                                 & {\color[HTML]{0606FF} \textbf{66.5}}  \\
    InternVL3-38B                     & 32.5              & 48.5           & 56.0            & {\color[HTML]{0606FF} 47.2}          &           & 35.4                                                          & 31.0                                                          & 39.4          & 66.6                                                         & 64.9                                                          & 38.0                                                           & 61.0                                                         & 77.3                                                          & {\color[HTML]{0606FF} 51.7}           \\
    $\textit{\method-S1}_\textnormal{InternVL3-38B}$                  & 39.0              & 68.0           & 76.8            & {\color[HTML]{0606FF} 64.6}          &           & 44.8                                                          & 47.0                                                          & 43.6          & 73.1                                                         & 68.6                                                          & 48.5                                                           & 71.2                                                         & 79.1                                                          & {\color[HTML]{0606FF} 59.5}           \\
    \rowcolor[HTML]{E7E6E6} 
    $\textit{\method-S1+S2}_\textnormal{InternVL3-38B}$           & \textbf{53.5}     & \textbf{85.2}  & \textbf{78.0}   & {\color[HTML]{0606FF} \textbf{77.4}} &           & \textbf{54.7}                                                 & \textbf{58.1}                                                 & \textbf{49.2} & \textbf{86.9}                                                & \textbf{80.4}                                                 & \textbf{49.1}                                                  & \textbf{79.6}                                                & \textbf{85.9}                                                 & {\color[HTML]{0606FF} \textbf{68.0}}  \\
    InternVL3-78B                     & 38.5              & 50.5           & 57.4            & {\color[HTML]{0606FF} 49.9}          &           & 54.6                                                          & 48.4                                                          & 50.3          & 77.7                                                         & 70.0                                                          & 44.8                                                           & 57.0                                                         & 76.6                                                          & {\color[HTML]{0606FF} 59.9}           \\
    $\textit{\method-S1}_\textnormal{InternVL3-78B}$                  & 43.5              & 69.0           & 77.2            & {\color[HTML]{0606FF} 66.1}          &           & 59.8                                                          & 53.1                                                          & 52.2          & 80.1                                                         & 72.5                                                          & 53.9                                                           & 65.1                                                         & 78.0                                                          & {\color[HTML]{0606FF} 64.3}           \\
    \rowcolor[HTML]{E7E6E6} 
    $\textit{\method-S1+S2}_\textnormal{InternVL3-78B}$           & \textbf{57.0}     & \textbf{86.2}  & \textbf{78.8}   & {\color[HTML]{FF0000} \textbf{78.9}} &           & \textbf{69.9}                                                 & \textbf{61.0}                                                 & \textbf{61.0} & \textbf{91.9}                                                & \textbf{88.6}                                                 & \textbf{54.8}                                                  & \textbf{75.3}                                                & \textbf{83.9}                                                 & {\color[HTML]{FF0000} \textbf{73.3}}  \\ \hline
    \end{tabular}
    \vspace{-1.5em}
\end{table*}
\noindent\textbf{Evaluation metric.} For multiple-choice questions, we use Accuracy, which is calculated based on exact matches between the model's predictions and the ground truth. For numerical-answer questions, we use Mean Relative Accuracy (MRA) introduced by~\cite{vsibench}, a metric that measures the closeness of the model's predictions to the ground truth values. "Avg." denotes the mean value of all subset task.

\noindent\textbf{Training.} Models marked with * are trained on our Large-Scale Curated Dataset, whereas those without the mark are trained exclusively on MindCube-only. For further details regarding the training dataset, please refer to the Supp. Mat..

\noindent\textbf{Hyper-parameters.} For \method trained on MindCube-only, in the stage 1, we set the MLP depth to 6, with the learning rate of 1e-4, latent size of 12, epoch of 10. In Eqn.~\ref{eq:sft}, the hyper-parameters $\lambda_{3D}, \lambda_{text}$ are uniformly set to 0.1 and 1, respectively. In the stage 2, we set the balancing coefficient of all three rewards to 1, with the learning rate of 1e-5, the rollout number of 8. Additional details are provided in the Supp. Mat..

\begin{table*}[t!]
    \caption{The evaluation of various baselines on the VSI-Bench~\cite{vsibench}, SPBench~\cite{li2025spatialladder}, CV-Bench~\cite{cvbench}, SPAR-Bench~\cite{spar}, ViewSpatial-Bench~\cite{viewspatial} and MMSI-Bench ~\cite{yang2025mmsi} datasets. [SI] denotes benchmarks with single image, whereas [MV] refers to multi-view images. The \textbf{best}-performing results under each base model are highlighted.}\label{table:baseline}
    \vspace{-0.8em}
    \centering
    \footnotesize
    \setlength{\tabcolsep}{0.7mm}
    \begin{tabular}{lccccccc}
    \hline
    \textbf{Method}  & \textbf{\begin{tabular}[c]{@{}c@{}}VSI-Bench~\cite{vsibench}\\ {[}MV{]}\end{tabular}} & \textbf{\begin{tabular}[c]{@{}c@{}}SPBench~\cite{li2025spatialladder}\\ {[}SI, MV{]}\end{tabular}} & \textbf{\begin{tabular}[c]{@{}c@{}}CV-Bench~\cite{cvbench}\\ {[}SI{]}\end{tabular}} & \textbf{\begin{tabular}[c]{@{}c@{}}SPAR-Bench~\cite{spar}\\ {[}SI, MV{]}\end{tabular}} & \textbf{\begin{tabular}[c]{@{}c@{}}ViewSpatial-Bench~\cite{viewspatial}\\ {[}SI, MV{]}\end{tabular}} & \textbf{\begin{tabular}[c]{@{}c@{}}MMSI-Bench~\cite{yang2025mmsi}\\ {[}MV{]}\end{tabular}} & {\color[HTML]{0606FF} \textbf{Avg.↑}} \\ \hline
    \multicolumn{8}{c}{\textit{Qwen2.5-VL-3B Based Spatial Models}}                                                                                                                                                                                                                                                                                                                                                                                                                                                             \\
    Qwen2.5-VL-3B~\cite{bai2025qwen2.5}    & 29.4                                                                  & 38.5                                                                    & 70.6                                                                 & 24.6                                                                       & 35.6                                                                              & 26.5                                                                   & {\color[HTML]{0606FF} 37.5}           \\
    Spatial-MLLM-4B~\cite{wu2025spatialmllm}  & 47.3                                                                  & 48.4                                                                    & 73.8                                                                 & 35.1                                                                       & 43.6                                                                              & 31.5                                                                   & {\color[HTML]{0606FF} 46.6}           \\
    SpatialLadder-3B~\cite{li2025spatialladder} & 45.7                                                                  & \textbf{70.6}                                                           & 73.7                                                                 & 34.4                                                                       & 44.2                                                                              & 29.2                                                                   & {\color[HTML]{0606FF} 49.6}           \\
    \rowcolor[HTML]{E7E6E6} 
    $\textit{\method-S1}_\textnormal{Qwen2.5-3B}*$                                                & 53.2                                                         & 54.8                                                                    & 74.5                                                       & 52.3                                                             & 59.5                                                                     & 37.7                                                          & {\color[HTML]{0606FF} 55.3}  \\

    \rowcolor[HTML]{E7E6E6} 
    $\textit{\method-S1+S2}_\textnormal{Qwen2.5-3B}*$        & \textbf{59.1}                                                         & 60.2                                                                    & \textbf{78.4}                                                        & \textbf{58.2}                                                              & \textbf{64.7}                                                                     & \textbf{41.9}                                                          & {\color[HTML]{0606FF} \textbf{60.4}}  \\ \hline
    \multicolumn{8}{c}{\textit{Qwen2.5-VL-7B Based Spatial Models}}                                                                                                                                                                                                                                                                                                                                                                                                                                                             \\
    Qwen2.5-VL-7B~\cite{bai2025qwen2.5}    & 35.8                                                                  & 42.9                                                                    & 73.0                                                                 & 30.2                                                                       & 37.9                                                                              & 26.9                                                                   & {\color[HTML]{0606FF} 41.1}           \\
    SpaceR-7B~\cite{ouyang2025spacer}        & 44.5                                                                  & 54.0                                                                    & 75.3                                                                 & 37.1                                                                       & 45.5                                                                              & 28.8                                                                   & {\color[HTML]{0606FF} 47.5}           \\
    VILASR-7B~\cite{vilasr}        & 45.4                                                                  & 53.9                                                                    & 77.1                                                                 & 37.8                                                                       & 46.1                                                                              & 30.2                                                                   & {\color[HTML]{0606FF} 48.4}           \\
    Video-R1~\cite{video-r1}         & 33.4                                                                  & 42.8                                                                    & 69.6                                                                 & 31.5                                                                       & 36.1                                                                              & 29.4                                                                   & {\color[HTML]{0606FF} 40.5}           \\
    \rowcolor[HTML]{E7E6E6}
    $\textit{\method-S1}_\textnormal{Qwen2.5-7B}*$                                                 & 57.3                                                         & 61.5                                                           & 77.9                                                        & 56.3                                                              & 61.7                                                                     & 41.5                                                          & {\color[HTML]{0606FF}59.4}  \\

    \rowcolor[HTML]{E7E6E6} 
    $\textit{\method-S1+S2}_\textnormal{Qwen2.5-7B}*$        & \textbf{63.7}                                                         & \textbf{68.3}                                                           & \textbf{81.1}                                                        & \textbf{63.3}                                                              & \textbf{68.6}                                                                     & \textbf{43.3}                                                          & {\color[HTML]{0606FF} \textbf{64.7}}  \\ \hline
    \end{tabular}
    \vspace{-1.8em}
\end{table*}
\subsection{Benchmarking Generalist VLMs}
In this section, we comprehensively investigate different training stages in \method across various generalist VLMs. We conduct experiments on MindCube-Tiny~\cite{mindcube} and Ego3D-Bench~\cite{ego3d}, both of which are designed to evaluate the spatial understanding ability from limited views. 

As shown in Tab.~\ref{table:mllm}, \method-full achieves consistent improvements over the generalist VLMs across all settings. On MindCube-Tiny, the overall performance gain ranges from \textbf{51.8\%} to \textbf{108.8\%}, while on Ego3D-Bench, the improvement spans \textbf{18.1\%} to \textbf{36.9\%}. Taking Qwen2.5-VL-3B as an example, \method boosts performance on MindCube-Tiny by \textbf{88.9\%} (62.7 vs. 33.2) after stage 1, and further improves by \textbf{19.9\%} (75.2 vs. 62.7) after stage 2. Similarly, on Ego3D-Bench, we observe a \textbf{19.3\%} improvement (46.7 vs. 39.1) after the stage 1 and an additional \textbf{8.8\%} gain (50.8 vs. 46.7) following the stage 2. 
Although the performance is slightly weaker on certain sub-tasks, e.g., Travel Time,
this can be attributed to the need for \emph{richer contextual information to align the normalized 3D representations with the real-world}. Remarkably, our model is trained without any Ego3D-specific data, yet it still achieves promising results on Ego3D-Bench, demonstrating strong cross-dataset generalization. \emph{This highlights that our "think with 3D" framework effectively enhances the model’s generalization capability across diverse spatial understanding scenarios.}
It is also worth noting that our best model, $\textit{\method-S1+S2}_\textnormal{Qwen2.5-72B}$, outperforms all other models, both open-source and closed-source, including the latest O3 model (\textbf{78.9} vs. 56.6 on MindCube-Tiny, \textbf{73.3} vs. 73.0 on Ego3D-Bench).

\subsection{Comparisons with Baselines}
We evaluate our method against several state-of-the-art (SOTA) approaches across a diverse set of categories. Additional details are provided in the Supp. Mat..

\noindent\textbf{Different Spatial Models.} 
As shown in Tab.~\ref{table:baseline}, we categorize the methods into two groups based on the types of base VLMs and then evaluate them across different benchmarks. 
For the Qwen2.5-VL-3B-based spatial models, \method surpasses the recent SOTA, SpatialLadder-3B, by \textbf{11.5\%} (55.3 vs. 49.6) in stage 1. This improvement is further enhanced to \textbf{21.8\%} (62.7 vs. 49.6) following stage 2.
When using the Qwen2.5-VL-7B model, our method achieves even more remarkable results. 
\method outperforms the SOTA VILASR-7B by \textbf{22.7\%} (59.4 vs. 48.4) in stage 1, and by \textbf{33.7\%} (64.7 vs. 48.4) in stage 2.
On the other hand, in contrast to methods that exhibit task-specific overfitting (e.g., SpatialLadder-3B on SPBench), \emph{\method demonstrates consistent improvement across all tasks, highlighting the robust spatial reasoning capability of our method.} 
Additionally, unlike models such as Video-R1, which struggle on single-view tasks (e.g., underperforming the base model on CV-Bench), \emph{our method demonstrates strong performance on both single-image and multi-view tasks.} This indicates that our 3D mental reasoning framework significantly enhances performance, even in single-image cases.

\noindent\textbf{Different Architectures and Parameter Scales.} 
Tab.~\ref{table:ego3d} compares our method with Ego3D-VLM on Ego3D-Bench across different model series and parameter scales. Although Ego3D-VLM constructs its cognitive map with the aid of external modules—specifically, a referring expression comprehension model (Grounding-DINO-Base~\cite{groundingdino}) and a depth estimator (Depth-Anything-V2-Metric-Large~\cite{yang2024depth})—\emph{our method, which does not rely on any extrinsic priors at inference, still achieves superior performance}. In particular, on Qwen2.5-VL-3B, \method yields a notable \textbf{14.4\%} improvement (50.8 vs. 44.4).
\begin{table}[t!]
    \caption{Performance on Ego3D-Bench (Accuracy Avg.) in comparison between \method and Ego3D-VLM, employing a series of VLMs with varying parameters. The \textbf{best} is highlighted.}\label{table:ego3d}
    \vspace{-0.8em}
    \centering
    \footnotesize
    \setlength{\tabcolsep}{1.1mm}
    \begin{tabular}{lccccccccc}
    \hline
                             & \multicolumn{4}{c}{\textbf{Qwen2.5-VL}}                                & \multicolumn{1}{l}{} & \multicolumn{4}{c}{\textbf{InternVL3}}                                 \\ \cline{2-5} \cline{7-10} 
    \multirow{-2}{*}{\textbf{Method}} & 3B            & 7B            & 32B           & 72B           &                      & 8B            & 14B           & 38B           & 78B           \\ \hline
    Ego3D-VLM~\cite{ego3d}                & 44.4          & 54.3          & 65.5          & 69.5          &                      & 60.1          & 66.1          & \textbf{68.0} & 71.8          \\
    \rowcolor[HTML]{E7E6E6} 
    \method                & \textbf{50.8} & \textbf{54.9} & \textbf{68.1} & \textbf{70.0} &                      & \textbf{60.4} & \textbf{66.5} & \textbf{68.0} & \textbf{73.3} \\ \hline
    \end{tabular}
    \vspace{-0.8em}
\end{table}
\begin{table}[t!]
    \caption{Results with Qwen2.5-VL-3B on MindCube-Tiny in terms of different training strategies. The \textbf{best} is highlighted.}\label{table:train}
    \vspace{-0.8em}
    \centering
    \footnotesize
    \setlength{\tabcolsep}{0.6mm}
    \begin{tabular}{lcccc}
    \hline
    \multicolumn{1}{c}{}                                  & \multicolumn{4}{c}{\textbf{MindCube-Tiny}}                                    \\ \cline{2-5} 
    \multicolumn{1}{l}{\multirow{-2}{*}{\textbf{Method}}} & \textbf{Rotation} & \textbf{Among} & \textbf{Around} & {\color[HTML]{0606FF} \textbf{Overall ↑}} \\ \hline
    raw-QA SFT                                            & 34.5              & 52.5           & 66.0            & {\color[HTML]{0606FF} 52.3}               \\
    CoT SFT                                               & 36.0              & 54.3           & 65.2            & {\color[HTML]{0606FF} 53.4}               \\
    Aug-CGMap-FFR-Out-SFT                             & \textbf{49.5}     & 52.5           & 66.4            & {\color[HTML]{0606FF} 55.2}               \\
    
    Plain-CGMap-FFR-Out-SFT                             & 47.5     & 62.3           & 67.6            & {\color[HTML]{0606FF} 60.8}               \\
    \rowcolor[HTML]{E7E6E6}
    $\textit{\method-S1}_\textnormal{Qwen2.5-3B}$                                      & 44.0              & \textbf{64.8}  & \textbf{72.4}   & {\color[HTML]{0606FF} \textbf{62.7}}      \\ \hline
    GRPO                                                  & 36.5              & 49.3           & 64.8            & {\color[HTML]{0606FF} 50.6}               \\
    CoT SFT + GRPO                                        & 36.5              & 55.2           & 65.6            & {\color[HTML]{0606FF} 54.1}               \\
    Aug-CGMap-FFR-Out-SFT+RL                            & 53.0              & 76.8           & 70.0            & {\color[HTML]{0606FF} 70.7}               \\
    Plain-CGMap-FFR-Out-SFT+RL                            & 48.0              & 79.2           & 68.4            & {\color[HTML]{0606FF} 70.7}               \\
    \rowcolor[HTML]{E7E6E6}
    $\textit{\method-S1+S2}_\textnormal{Qwen2.5-3B}$                               & \textbf{55.5}     & \textbf{81.8}  & \textbf{75.2}   & {\color[HTML]{0606FF} \textbf{75.2}}      \\ \hline
    \end{tabular}
    \vspace{-1.0em}
\end{table}

\subsection{Training Strategies}
To further demonstrate the effectiveness of our training paradigm, we compare \method against several representative training strategies as shown in Tab.~\ref{table:train}. Among them, Aug-CGMap-FFR-Out and Plain-CGMap-FFR-Out serve as SOTA baselines introduced in~\cite{mindcube}. Specifically, Aug-CGMap-FFR-Out performs reasoning with the augmented cognitive map (camera-view information included), whereas Aug-CGMap-FFR-Out relies solely on plain cognitive maps without augmentation.

Under supervised training, our method surpasses raw-QA SFT, CoT SFT, and even the cognitive-map-based SFT proposed in~\cite{mindcube} by a margin of \textbf{3.1\%} (62.7 vs. 60.8). The relatively smaller improvement observed in the rotation sub-category can be attributed to its requirement for dynamic spatial imagination. \emph{Since our "think with 3D" supervised framework primarily targets static spatial understanding, the RL stage further enhances its dynamic capability by optimizing whole reasoning trajectories.} That is, through outcome-based RL, \method progressively refines the 3D latents across rollouts, achieving additional gains in both zero-RL and SFT-then-RL settings. Furthermore, \method achieves a \textbf{6.4\%} improvement over the cognitive-map-based SFT-then-RL baseline (75.2 vs. 70.7), demonstrating its superior capability in integrating spatial reasoning with reinforcement learning.

\subsection{Visualization}
\begin{figure}[]
    \centering
    \includegraphics[width=0.48\textwidth]{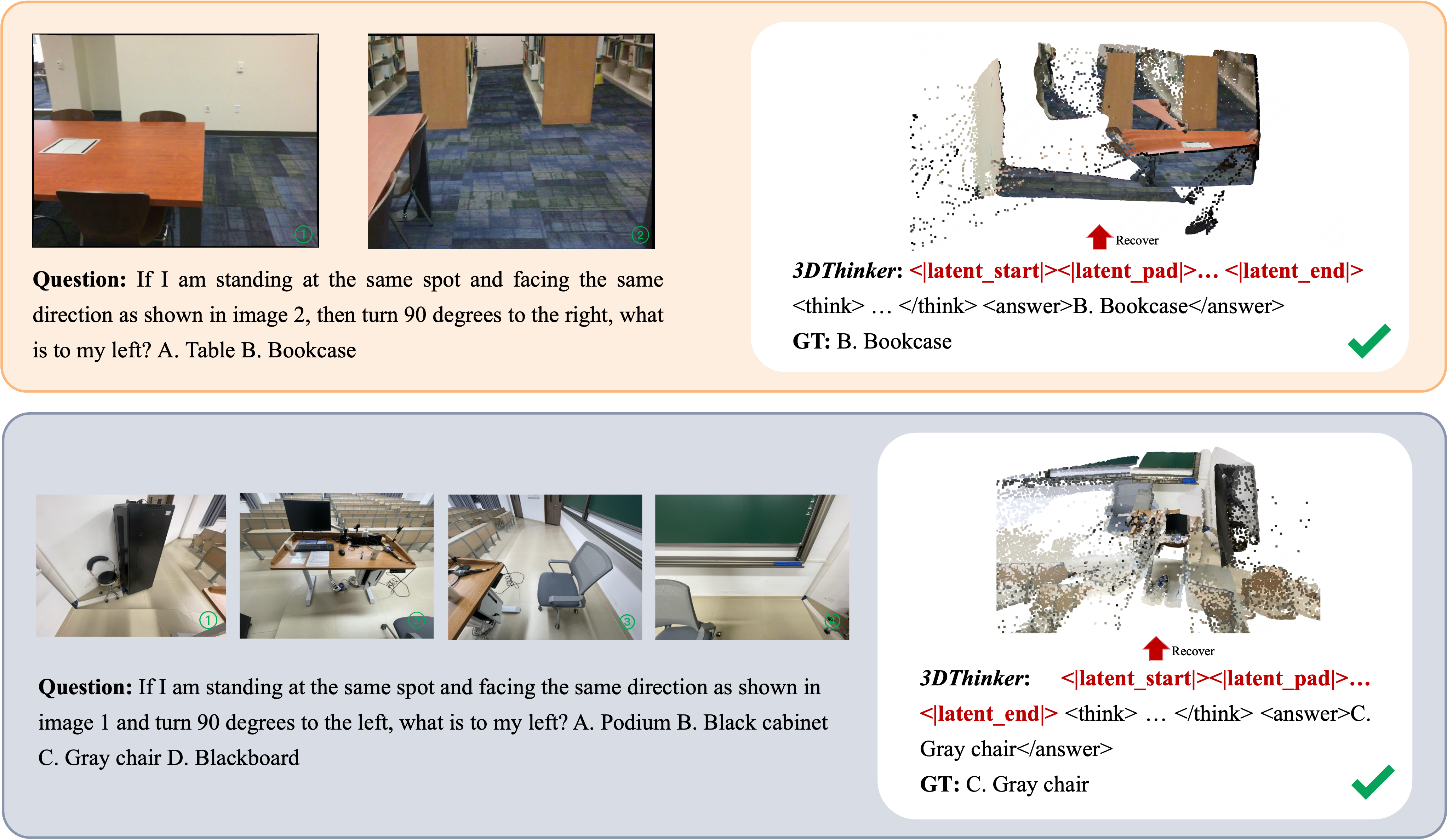}
    \vspace{-1.8em}
    \caption{The reasoning process for different cases is presented, along with the visualization of the 3D latent representations.}\label{fig:vis}
    \vspace{-0.8em}
\end{figure}
We visualize the results in Fig.~\ref{fig:vis}. During inference, we extract the last layer hidden states corresponding to the 3D special tokens. These 3D latents are projected into the VGGT feature space via the projector illustrated in Fig.~\ref{fig:projector}. The projected features are subsequently processed by the DPT~\cite{dpt} of VGGT to generate point clouds. As shown in Fig.~\ref{fig:vis}, the reconstructed point clouds roughly depict the underlying scene, \emph{where the clearer regions are typically correlated with prompt-relevant objects}. This observation indicates that the 3D latents effectively encode the mental scene guided by the prompt intent. After reasoning with 3D mentaling, all two examples yield correct answers. Additional results are provided in the Supp. Mat..

\subsection{Ablation Study}
\noindent\textbf{Different 3D Latent Size.} 
In Tab.~\ref{table:ablation_latent}, we ablate the effect of different latent sizes on the results. The results indicate that the optimal performance is achieved with the latent size of about 12. \emph{This is because a smaller latent size limits the model's representational capacity, while a larger latent size can compromise the model's natural expressive ability}, leading to repetitive {\small \texttt{<|latent\_start|>}} outputs that fail to yield the final answer.
\begin{table}[t!]
    \caption{Ablation of different 3D latent size on MindCube-Tiny in terms of $\textit{\method-S1}_\textnormal{Qwen2.5-3B}$.}\label{table:ablation_latent}
    \vspace{-0.8em}
    \centering
    \footnotesize
    \begin{tabular}{lcc
    >{\columncolor[HTML]{E7E6E6}}c ccc}
    \hline
    \textbf{Latent Size}               & 4    & 8    & 12            & 16   & 32   & 64   \\ \hline
    \textbf{Accuracy} & 60.2 & 60.6 & \textbf{62.7} & 59.9 & 25.1 & 15.5 \\ \hline
    \end{tabular}
    \vspace{-1.8em}
\end{table}

\noindent\textbf{Different Designs.} 
As shown in Tab.~\ref{table:ablation_module}, we first conduct an ablation study on the placement of the 3D special tokens. Beyond the approach in Sec.~\ref{sec:rl}, where the special tokens is positioned at the beginning (before {\small \texttt{<think>}}), we also explore placing it between the {\small \texttt{<think>}} and {\small \texttt{</think>}}, as well as at the end (after {\small \texttt{</answer>}}). We observe that placing the 3D tokens in the middle disrupts natural language coherence: \emph{the 3D latent can resemble certain character features, leading to garbled text and premature output termination}. This results in a significant performance drop (75.2 vs. 42.0). In contrast, positioning the 3D tokens at the beginning or end—where it is isolated from natural text—yields significantly better performance.
\begin{table}[t!]
    \caption{Ablation of different designs including 3D special token position (Token Pos.), projector and rewards in terms of $\textit{\method-S1+S2}_\textnormal{Qwen2.5-3B}$ on MindCube-Tiny.}\label{table:ablation_module}
    \vspace{-0.8em}
    \centering
    \scriptsize
    \setlength{\tabcolsep}{0.7mm}
    \begin{tabular}{lcclclcccl
    >{\columncolor[HTML]{E7E6E6}}c }
    \hline
                                      & \multicolumn{2}{c}{\textbf{Token Pos.}} & \multicolumn{1}{c}{} & \textbf{Projector} &  & \multicolumn{3}{c}{\textbf{Rewards}}   & \multicolumn{1}{c}{} &               \\ \cline{2-3} \cline{5-5} \cline{7-9}
    \multirow{-2}{*}{\textbf{Method}} & Middle           & End           &                      & VGGT-to-VLM   &  & w/o $r_{\text{format}}$ & w/o $r_{\text{ans}}$ & w/o $r_{\text{3D}}$ &                      & Full     \\ \hline
    \textbf{Accuracy}            & 42.0                & 74.3             &                      & 74.1               &  & 74.8 & 64.2       & 68.3      &                      & \textbf{75.2} \\ \hline
    \end{tabular}
    \vspace{-2.5em}
\end{table}

We also examine two potential projector configurations. The first maps the last layer hidden state of the VLM to the VGGT space (shown in Fig.~\ref{fig:projector}), \emph{allowing the VLM features to be explicitly converted into 3D representations (e.g., point clouds) via the projector}. The alternative compresses VGGT features directly into the VLM space (e.g., via adaptive average pooling), but this approach is unrecoverable to 3D representations. Given the interpretability, visualizability, and better performance (75.2 vs. 74.1) of the first approach, we adopt it as our projector strategy.

Finally, we ablate the three rewards used in stage 2. Among them, the formatting requirement has minimal impact. In contrast, removing 3D alignment leads to a substantial performance drop (75.2 vs. 68.3) due to \emph{the absence of stable constraints on the 3D latent}. The final answer reward is also critical (75.2 vs. 64.2), serving as \emph{the sole ground-truth supervision signal} and guiding optimization of each token across the entire rollout.

\section{Conclusion}
\label{sec:conclusion}
In this paper, we propose \method, a framework for VLM to think with 3D spatial mentaling. Unlike recent methods that rely solely on pure text or 2D visual cues for reasoning, \method leverages geometric information embedded in images during the reasoning process for the first time. Additionally, our method does not rely on dense annotations or other external priors. To enable thinking with 3D spatial mentaling, we introduce a two-stage training scheme. Stage 1 distills geometric features from a pretrained 3D model to warm up. Stage 2 optimizes the entire reasoning trajectory while maintaining 3D visual alignment based on the outcome signal. Experimental results show that our method outperforms previous methods across various benchmarks, establishing a solid foundation for future exploration.

\section{Acknowledgments} 
This work was supported by the National Natural Science Foundation of China under contract No. 62171256.
{
    \small
    \bibliographystyle{ieeenat_fullname}
    \bibliography{main}
}
\label{sec:appendix}
\clearpage
\setcounter{page}{1}
\maketitlesupplementary
\label{sec:appendix}
In this supplementary material, we provide more technical details and experimental results, including 1) Detailed descriptions of training dataset and benchmarks in Sec.~\ref{sec:dataset}; 2) Additional results including benchmarking Qwen3-VL and LLaVA-OneVision-1.5, additional baselines, general image understanding, prompt-relevant region, as well as ablation studies on projection methods and 3D loss in Sec.~\ref{sec:results}; 3) Training curve for two stages of \method in Sec.~\ref{sec:curve}; 4) Details of the dataset generation process in Sec.~\ref{sec:instruction}; 5) Explanation of the claim in Sec.~\ref{sec:claim}; 6) Cost in Sec.~\ref{sec:cost}; and 7) More "think with 3D" visualization including representative and failure cases in Sec.~\ref{sec:visualization}.

\section{Dataset}\label{sec:dataset}
\subsection{Training Dataset}
\noindent\textbf{MindCube-Only Training:} Following~\cite{mindcube}, we utilize a subset of 10,000 question-images-answer pairs for the training phases in Tab.1 of the main text. That is, in stage 1, we use the 10,000 samples to construct the CoT data for supervised training. In stage 2, we utilize only the question–image–answer pairs of these 10,000 samples for reinforced training. \emph{Notably, no annotated cognitive maps are used in any stage of our training.} 

\noindent\textbf{Large-Scale Curated Dataset Training:} We further select data from MM-Spatial~\cite{mmspatial}, SPAR~\cite{spar}, MindCube~\cite{mindcube}, and SpatialLadder~\cite{li2025spatialladder} to construct a larger-scale dataset. We utilized Gemini 2.5 Pro to evaluate the data quality and balanced the distribution between single-image and multi-image categories, ultimately retaining 237K samples. The methodology for constructing CoT data from standard QA pairs is detailed in Sec. 3.1 in the main text and Sec.~\ref{sec:instruction}. Similarly, we employed the complete CoT dataset in stage 1, whereas stage 2 solely utilized the question-image-answer pairs. For the training process, stage 1 was trained on the entire dataset for one epoch, while stage 2 was trained on a sampled subset of 57K instances.

\subsection{Benchmarks}
\noindent\textbf{MindCube-Tiny}~\cite{mindcube} contains 1,050 data, serves as the benchmark for evaluation of spatial understanding from limited views. The dataset includes three types of camera movements: Rotation (the camera remains stationary while rotating to observe the surroundings), Around (the camera moves in a circular path around the evaluated objects), and Among (the camera moves in a circular path among the evaluated objects).

\noindent\textbf{Ego3D-Bench}~\cite{ego3d} is a comprehensive benchmark consisting of over 8,600 question–answer pairs, specifically designed to evaluate the spatial reasoning capabilities of model in ego-centric, multi-view outdoor environments. The dataset encompasses five tasks—Absolute Distance Measurement (Dist.), Relative Distance Measurement (Rel.), Localization (Loc.), Motion Reasoning (Mot.), and Travel Time (Time), each formulated in both ego-centric (Ego) and object-centric (Obj.) settings.

\noindent\textbf{VSI-Bench}~\cite{vsibench} consists of over 5,000 question–answer pairs derived from 288 real-world videos. These videos encompass a wide variety of environments—including residential areas, professional settings (e.g., offices and laboratories), and industrial sites (e.g., factories), and span multiple geographic regions.

\noindent\textbf{SPBench}~\cite{li2025spatialladder} comprises two different tasks: SPBench-SI and SPBench-MV, corresponding to the single-image and multi-view types, respectively. SPBench-SI includes four categories: absolute distance, object size, relative distance, and relative direction, containing a total of 1,009 samples. SPBench-MV extends the evaluation by incorporating object counting tasks, comprising 319 samples.

\noindent\textbf{CV-Bench}~\cite{cvbench} contains 2,638 manually inspected examples, each formulated as a natural-language question to assess the fundamental 2D and 3D understanding capabilities of VLMs. The 2D understanding tasks focus on spatial relationships and object counting, while the 3D understanding tasks involve depth ordering and relative distance.

\noindent\textbf{SPAR-Bench}~\cite{spar} covers 20 diverse spatial tasks—including depth estimation, distance measurement, spatial relations, etc. The benchmark spans single-view, multi-view, and video settings, comprising a total of 7,207 manually verified question–answer pairs.

\noindent\textbf{ViewSpatial-Bench}~\cite{viewspatial} focuses on reasoning from alternative spatial frames of reference, requiring models to adopt another entity’s viewpoint. It consists of over 5,700 question–answer pairs derived from more than 1,000 3D scenes. This benchmark evaluates the spatial localization capabilities, specifically assessing both egocentric (camera-centered) and allocentric (human-centered) viewpoints across five distinct task types.

\noindent\textbf{MMSI-Bench}~\cite{yang2025mmsi} comprises 1,000  multiple-choice questions from over 120,000 images. The benchmark focuses on positional relationships (six pairwise combinations of camera, object, and region), attributes (measurement and appearance), and motion (camera and object).

\section{Additional Results}\label{sec:results}
\subsection{Benchmarking Qwen3-VL and LLaVA-OneVision-1.5}
\begin{table*}[t!]
    \caption{Accuracy comparison between \textbf{LATEST} Qwen3-VL (under varying parameter scales and architectures) and \method on different benchmarks, with our training conducted on stage 1 (S1) and on both stage 1 and stage 2 (S1 + S2). The best results achieved based on different VLMs are \textbf{bolded}. The overall/average results of each model are highlighted in \textcolor{blue}{blue}.}\label{table:sup_qwen}
    \vspace{-0.8em}
    \centering
    \footnotesize
    \setlength{\tabcolsep}{1.4mm}
    \begin{tabular}{lcccclccccccccc}
    \hline
                                      & \multicolumn{4}{c}{\textbf{MindCube-Tiny~\cite{mindcube}}}                                                       & \textbf{}                     & \multicolumn{9}{c}{\textbf{Ego3D-Bench~\cite{ego3d}}}                                                                                                                                                                                                                                                                                                                                                                                                                                                                           \\ \cline{2-5} \cline{7-15} 
    \multirow{-2}{*}{\textbf{Method}} & \textbf{Rotation} & \textbf{Among} & \textbf{Around} & {\color[HTML]{0606FF} \textbf{Overall ↑}} & \multicolumn{1}{c}{\textbf{}} & \textbf{\begin{tabular}[c]{@{}c@{}}Ego \\ Dist.\end{tabular}} & \textbf{\begin{tabular}[c]{@{}c@{}}Obj.\\ Dist.\end{tabular}} & \textbf{Loc.} & \textbf{\begin{tabular}[c]{@{}c@{}}Ego \\ Mot.\end{tabular}} & \textbf{\begin{tabular}[c]{@{}c@{}}Obj. \\ Mot.\end{tabular}} & \textbf{\begin{tabular}[c]{@{}c@{}}Travel\\ Time\end{tabular}} & \textbf{\begin{tabular}[c]{@{}c@{}}Ego\\ Rel.\end{tabular}} & \textbf{\begin{tabular}[c]{@{}c@{}}Obj.\\ Rel.\end{tabular}} & {\color[HTML]{0606FF} \textbf{Avg.↑}} \\ \hline
    Qwen3-VL-2B                       & 29.0              & 32.2           & 46.8            & {\color[HTML]{0606FF} 35.0}               &                               & 24.7                                                          & 23.9                                                          & 30.4          & 58.7                                                         & 56.4                                                          & 32.8                                                           & 57.4                                                        & 61.1                                                         & {\color[HTML]{0606FF} 43.2}           \\
    $\textit{\method-S1}_\textnormal{Qwen3-2B}$                  & 40.5              & 67.5           & 77.2            & {\color[HTML]{0606FF} 64.7}               &                               & 45.9                                                          & 41.6                                                          & 36.0          & 60.6                                                         & 59.2                                                          & 37.2                                                           & 62.0                                                        & 72.1                                                         & {\color[HTML]{0606FF} 51.8}           \\
    \rowcolor[HTML]{E7E6E6} 
    $\textit{\method-S1+S2}_\textnormal{Qwen3-2B}$           & \textbf{53.0}     & \textbf{83.2}  & \textbf{78.4}   & {\color[HTML]{0606FF} \textbf{76.2}}      &                               & \textbf{53.1}                                                 & \textbf{51.3}                                                 & \textbf{36.4} & \textbf{61.4}                                                & \textbf{62.8}                                                 & \textbf{38.2}                                                  & \textbf{65.3}                                               & \textbf{80.5}                                                & {\color[HTML]{0606FF} \textbf{56.1}}  \\
    Qwen3-VL-4B                       & 34.0              & 18.7           & 37.2            & {\color[HTML]{0606FF} 26.0}               &                               & 41.3                                                          & 40.9                                                          & 27.9          & 56.9                                                         & 56.6                                                          & 37.8                                                           & 60.1                                                        & 68.2                                                         & {\color[HTML]{0606FF} 48.7}           \\
    $\textit{\method-S1}_\textnormal{Qwen3-4B}$                  & 43.0              & 65.3           & 76.0            & {\color[HTML]{0606FF} 63.6}               &                               & 54.0                                                          & 49.9                                                          & 35.1          & 59.3                                                         & 60.8                                                          & \textbf{40.8}                                                  & 64.9                                                        & 76.1                                                         & {\color[HTML]{0606FF} 55.1}           \\
    \rowcolor[HTML]{E7E6E6} 
    $\textit{\method-S1+S2}_\textnormal{Qwen3-4B}$           & \textbf{55.0}     & \textbf{81.7}  & \textbf{78.4}   & {\color[HTML]{0606FF} \textbf{75.8}}      &                               & \textbf{62.3}                                                 & \textbf{56.5}                                                 & \textbf{35.5} & \textbf{61.4}                                                & \textbf{64.1}                                                 & 40.4                                                           & \textbf{68.3}                                               & \textbf{84.0}                                                & {\color[HTML]{0606FF} \textbf{59.1}}  \\
    Qwen3-VL-8B                       & 31.5              & 31.0           & 41.2            & {\color[HTML]{0606FF} 33.5}               &                               & 38.9                                                          & 32.6                                                          & 31.6          & 45.3                                                         & 56.8                                                          & 40.8                                                           & 62.4                                                        & 69.3                                                         & {\color[HTML]{0606FF} 47.2}           \\
    $\textit{\method-S1}_\textnormal{Qwen3-8B}$                  & 42.0              & 67.8           & 76.8            & {\color[HTML]{0606FF} 65.0}               &                               & 52.1                                                          & 45.9                                                          & 37.1          & 51.5                                                         & 61.8                                                          & \textbf{43.2}                                                  & 67.8                                                        & 78.2                                                         & {\color[HTML]{0606FF} 54.7}           \\
    \rowcolor[HTML]{E7E6E6} 
    $\textit{\method-S1+S2}_\textnormal{Qwen3-8B}$           & \textbf{56.5}     & \textbf{82.5}  & \textbf{78.8}   & {\color[HTML]{0606FF} \textbf{76.7}}      &                               & \textbf{63.0}                                                 & \textbf{56.8}                                                 & \textbf{38.1} & \textbf{55.5}                                                & \textbf{66.2}                                                 & \textbf{43.2}                                                  & \textbf{71.1}                                               & \textbf{86.1}                                                & {\color[HTML]{0606FF} \textbf{60.0}}  \\
    Qwen3-VL-32B                      & 29.5              & 33.5           & 38.0            & {\color[HTML]{0606FF} 33.8}               &                               & 51.5                                                          & 40.7                                                          & 40.9          & 76.4                                                         & 58.2                                                          & 40.4                                                           & 61.5                                                        & 81.4                                                         & {\color[HTML]{0606FF} 56.4}           \\
    $\textit{\method-S1}_\textnormal{Qwen3-32B}$                  & 41.0              & 69.5           & 77.6            & {\color[HTML]{0606FF} 66.1}               &                               & 60.4                                                          & 51.0                                                          & 48.3          & 78.6                                                         & 67.9                                                          & \textbf{43.2}                                                  & 68.1                                                        & 87.3                                                         & {\color[HTML]{0606FF} 63.1}           \\
    \rowcolor[HTML]{E7E6E6} 
    $\textit{\method-S1+S2}_\textnormal{Qwen3-32B}$           & \textbf{55.0}     & \textbf{84.2}  & \textbf{79.6}   & {\color[HTML]{0606FF} \textbf{77.5}}      &                               & \textbf{69.3}                                                 & \textbf{62.2}                                                 & \textbf{48.7} & \textbf{85.1}                                                & \textbf{73.5}                                                 & 43.0                                                           & \textbf{73.1}                                               & \textbf{90.3}                                                & {\color[HTML]{0606FF} \textbf{68.2}}  \\
    Qwen3-VL-30B-A3B                  & 37.0              & 43.9           & 45.6            & {\color[HTML]{0606FF} 42.9}               &                               & 49.1                                                          & 45.8                                                          & 28.4          & 64.5                                                         & 60.0                                                          & 34.7                                                           & 63.3                                                        & 73.3                                                         & {\color[HTML]{0606FF} 52.4}           \\
    $\textit{\method-S1}_\textnormal{Qwen3-30B-A3B}$                  & 46.5              & 72.8           & 82.0            & {\color[HTML]{0606FF} 70.0}               &                               & 60.1                                                          & 54.4                                                          & 38.3          & 72.2                                                         & 70.3                                                          & \textbf{38.0}                                                  & 71.0                                                        & 80.1                                                         & {\color[HTML]{0606FF} 60.6}           \\
    \rowcolor[HTML]{E7E6E6} 
    $\textit{\method-S1+S2}_\textnormal{Qwen3-30B-A3B}$           & \textbf{60.0}     & \textbf{86.7}  & \textbf{83.2}   & {\color[HTML]{0606FF} \textbf{80.8}}      &                               & \textbf{69.4}                                                 & \textbf{65.1}                                                 & \textbf{39.1} & \textbf{79.3}                                                & \textbf{75.1}                                                 & 37.6                                                           & \textbf{76.3}                                               & \textbf{88.0}                                                & {\color[HTML]{0606FF} \textbf{66.2}}  \\ \hline
    \end{tabular}
\end{table*}
\begin{table*}[t!]
    \caption{Accuracy comparison of generalist VLMs and our method (\method) on MindCube-Tiny and Ego3D-Bench, with our training conducted on stage 1 (S1) and on both stage 1 and stage 2 (S1 + S2). The best results achieved based on different VLMs are \textbf{bolded}. The overall/average results of each model are highlighted in \textcolor{blue}{blue}, with the best results among all models highlighted in \textbf{\textcolor{red}{red}}.}\label{table:llavaov}
    \vspace{-0.8em}
    \centering
    \footnotesize
    \setlength{\tabcolsep}{1.4mm}
    \begin{tabular}{lcccccccccccccc}
    \hline
                                      & \multicolumn{4}{c}{\textbf{MindCube-Tiny}}                               & \textbf{} & \multicolumn{9}{c}{\textbf{Ego3D-Bench}}                                                                                                                                                                                                                                                                                                                                                                                                                                                          \\ \cline{2-5} \cline{7-15} 
    \multirow{-2}{*}{\textbf{Method}} & \textbf{Rotation} & \textbf{Among} & \textbf{Around} & {\color[HTML]{0606FF} \textbf{Overall ↑}}     & \textbf{} & \textbf{\begin{tabular}[c]{@{}c@{}}Ego \\ Dist.\end{tabular}} & \textbf{\begin{tabular}[c]{@{}c@{}}Obj.\\ Dist.\end{tabular}} & \textbf{Loc.} & \textbf{\begin{tabular}[c]{@{}c@{}}Ego \\ Mot.\end{tabular}} & \textbf{\begin{tabular}[c]{@{}c@{}}Obj. \\ Mot.\end{tabular}} & \textbf{\begin{tabular}[c]{@{}c@{}}Travel\\ Time\end{tabular}} & \textbf{\begin{tabular}[c]{@{}c@{}}Ego \\ Rel.\end{tabular}} & \textbf{\begin{tabular}[c]{@{}c@{}}Obj. \\ Rel.\end{tabular}} & {\color[HTML]{0606FF} \textbf{Avg.↑}} \\ \hline
    \multicolumn{15}{c}{{\color[HTML]{000000} \textit{LLaVA-OneVision-1.5   Family}~\cite{an2025llavao1.5}}}                                                                                                                                                                                                                                                                                                                                                                                                                                                                                                                                                                                  \\
    LLaVA-OneVision-1.5-4B            & 33.5              & 38.0           & 49.2            & {\color[HTML]{0606FF} 39.8}          &           & 39.7                                                          & 37.1                                                          & 29.2          & 51.4                                                         & 51.8                                                          & \textbf{34.1}                                                  & 52.4                                                         & 73.5                                                          & {\color[HTML]{0606FF} 46.2}           \\
    $\textit{\method-S1}_\textnormal{LLaVA-O-1.5-4B}$                  & 41.5              & 59.8           & \textbf{66.0}   & {\color[HTML]{0606FF} 57.8}          &           & 40.0                                                          & 39.1                                                          & 33.1          & 51.1                                                         & \textbf{52.6}                                                 & 30.9                                                           & 58.6                                                         & \textbf{73.8}                                                 & {\color[HTML]{0606FF} 47.4}           \\
    \rowcolor[HTML]{E7E6E6} 
    $\textit{\method-S1+S2}_\textnormal{LLaVA-O-1.5-4B}$           & \textbf{48.0}     & \textbf{67.5}  & 65.2            & {\color[HTML]{0606FF} \textbf{63.2}} &           & \textbf{40.2}                                                 & \textbf{39.9}                                                 & \textbf{34.2} & \textbf{51.9}                                                & 52.3                                                          & 30.8                                                           & \textbf{61.8}                                                & \textbf{73.8}                                                 & {\color[HTML]{0606FF} \textbf{48.1}}  \\
    LLaVA-OneVision-1.5-8B            & 34.5              & 34.7           & 48.4            & {\color[HTML]{0606FF} 37.9}          &           & 30.3                                                          & 36.6                                                          & 34.3          & 44.9                                                         & 51.9                                                          & \textbf{36.9}                                                  & 53.4                                                         & 74.4                                                          & {\color[HTML]{0606FF} 45.3}           \\
    $\textit{\method-S1}_\textnormal{LLaVA-O-1.5-8B}$                  & 43.0              & 57.8           & \textbf{64.8}   & {\color[HTML]{0606FF} 56.7}          &           & 35.1                                                          & 39.0                                                          & 36.1          & 44.9                                                         & 53.2                                                          & 31.9                                                           & 61.0                                                         & 73.8                                                          & {\color[HTML]{0606FF} 46.9}           \\
    \rowcolor[HTML]{E7E6E6} 
    $\textit{\method-S1+S2}_\textnormal{LLaVA-O-1.5-8B}$           & \textbf{49.0}     & \textbf{68.2}  & \textbf{64.8}   & {\color[HTML]{0606FF} \textbf{63.7}} &           & \textbf{36.5}                                                 & \textbf{41.5}                                                 & \textbf{37.0} & \textbf{46.2}                                                & \textbf{53.3}                                                 & 32.8                                                           & \textbf{64.9}                                                & \textbf{77.2}                                                 & {\color[HTML]{0606FF} \textbf{48.7}}  \\ \hline
    \end{tabular}
\end{table*}
We further conducted experiments on the Qwen3-VL and LLaVA-OneVision-1.5. Note that Qwen3-VL adopts {\small \texttt{<think>}} as a predefined special token, thus we have modified to {\small \texttt{<thinking>}} to avoid potential conflicts. On the other hand, \emph{we modify the stage 2 from token-level GRPO~\cite{rela:deepseekr1} to sequence-level GSPO~\cite{gspo} for all Mixture-of-Experts(MoE)-based~\cite{jacobs1991adaptive} models} to maintain stable training. 

\noindent\textbf{Qwen3-VL:} Tab.~\ref{table:sup_qwen} shows that our method consistently boosts the spatial understanding capability of Qwen3-VL. On the MindCube-Tiny~\cite{mindcube} dataset, \method achieves a maximum improvement of \textbf{191.5\%} (75.8 vs. 25.0), elevating the VLM from random guessing to highly accurate understanding on spatial reasoning tasks. On the Ego3D-Bench~\cite{ego3d} dataset, \method yields up to a \textbf{29.9\%} improvement (56.1 vs. 43.2), demonstrating strong generalization ability. Moreover, we validated our method on Qwen3-VL-30B-A3B sparse model as well, confirming the effectiveness of our method across both dense and sparse architectures.

\noindent\textbf{LLaVA-OneVision-1.5:} As shown in Tab.~\ref{table:llavaov}, we further extended our training to another foundation model. Specifically, on LLaVA-OneVision-1.5, \method consistently yielded significant performance gains. Notably, it achieved a \textbf{68.07\% } (63.7 vs. 37.9) improvement on MindCube-Tiny and a \textbf{7.5\%} (48.7 vs. 45.3) increase on Ego3D-Bench.

\subsection{Additional Baselines}
As shown in Tab.~\ref{table:3drs}, \method \emph{consistently outperforms previous methods}, surpassing the SOTA by \textbf{8.79\%} (65.6 vs. 60.3). This further highlights the effectiveness of incorporating 3D mentaling for spatial reasoning.
\begin{table}[t!]
    \caption{Performance comparison of all baselines on 3DSRBench.}\label{table:3drs}
    \vspace{-0.8em}
    \centering
    \footnotesize
    \setlength{\tabcolsep}{1.5mm}
    \begin{tabular}{ccc
    >{\columncolor[HTML]{E7E6E6}}c }
    \hline
    SpatialReasoner & SpatialReasoner-R1 & SpatialThinker & \method*     \\ \hline
    60.3            & 55.7               & 56.4           & \textbf{65.6} \\ \hline
    \end{tabular}
\end{table}

\subsection{General Image Understanding}
As shown in Tab.~\ref{table:general}, our method largely maintains, and on some benchmarks even substantially improves. For example, on POPE \method outperforms the base model by \textbf{2.91\%} (88.4 vs. 85.9). On MME, which predominantly consists of 2D image understanding tasks (e.g., OCR, numerical reasoning), our results are still comparable to the base model. These results suggest that \method effectively transfers the base model’s 2D image understanding abilities and \emph{exhibits strong robustness}.
\begin{table}[t!]
    \caption{Results on general VLM benchmarks.}\label{table:general}
    \vspace{-0.8em}
    \centering
    \footnotesize
    \setlength{\tabcolsep}{2.8mm}
    \begin{tabular}{lcccc}
    \hline
    \textbf{Method}     & \textbf{MME\textsuperscript{P}} & \textbf{MME\textsuperscript{C}} & \textbf{POPE}  & \textbf{SEED-I}        \\ \hline
    Qwen2.5-VL-7B & 1670                   & \textbf{623}           & 85.9 & 77.0 \\
    \rowcolor[HTML]{E7E6E6} 
    \method*  & \textbf{1677}          & 610                    & \textbf{88.4} & \textbf{78.9} \\ \hline
    \end{tabular}
\end{table}

\subsection{Prompt-Relevant Region}
\begin{figure}[]
    \centering
    \includegraphics[width=0.4\textwidth]{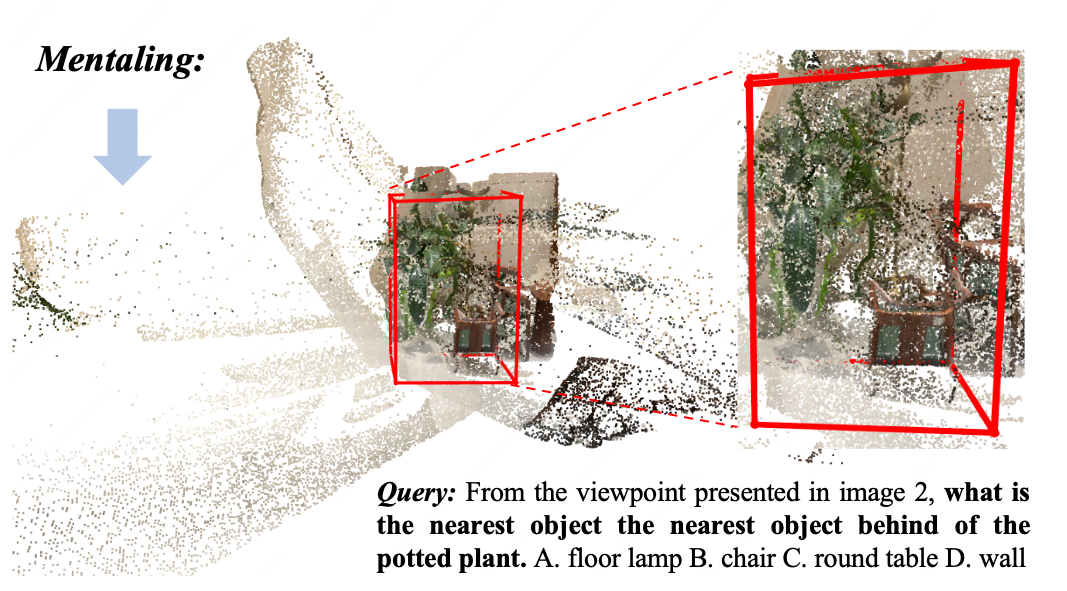}
    \caption{Density calculation for the Mentaled point cloud.}\label{fig:density}
\end{figure}

As shown in Fig.~\ref{fig:density}, we provide a quantitative result to substantiate the claim of ``prompt-relevant''. Specifically, we compute the point cloud density within the red 3D bounding box and compare it with the overall point cloud density. The resulting densities are \textbf{196494.67} vs. 1332.54 $\text{points/unit}^3$.

\subsection{Ablation on Projection Methods}
\begin{figure}[]
    \centering
    \includegraphics[width=0.5\textwidth]{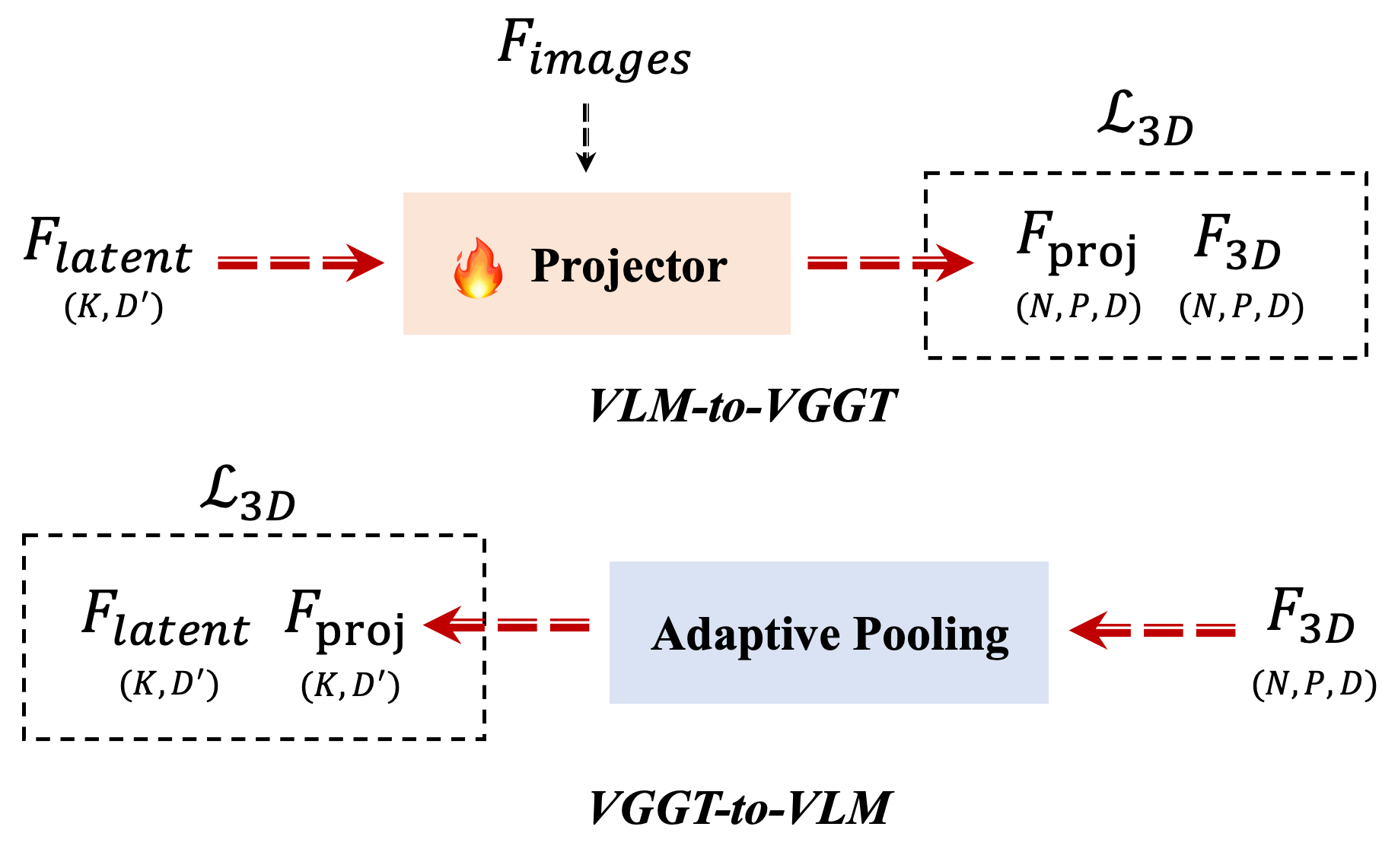}
    \vspace{-0.8em}
    \caption{Illustration of two methods for projection. Above: project the latents generated by VLM into the feature space of the 3D foundation model; Below: project the features of the 3D foundation model into the feature space of the VLM.}\label{fig:proj}
\end{figure}
As illustrated in Fig.~\ref{fig:proj}, we present a more detailed comparison of the two feature alignment strategies. The above method requires training the projector, whereas the below one can be implemented directly via pooling~\cite{3drs} without any training. As discussed in Sec. 4.5, the two methods achieve comparable performance (74.1 vs. \textbf{75.2}). However, the above method not only maintains slightly better performance, but also \emph{enables the recovery of 3D representations from the VLM latent space}. Therefore, we adopt the VLM-to-VGGT projection in our framework.

\subsection{Ablation on 3D Loss}
\begin{table}[]
    \caption{Ablation study on 3D alignment loss  on MindCube-Tiny in terms of Qwen2.5-VL-3B.}\label{table:3dloss}
    \vspace{-0.8em}
    \centering
    \footnotesize
    \begin{tabular}{lcccc}
    \hline
    \multicolumn{1}{c}{}                                  & \multicolumn{4}{c}{\textbf{MindCube-Tiny}}                                    \\ \cline{2-5} 
    \multicolumn{1}{l}{\multirow{-2}{*}{\textbf{Method}}} & \textbf{Rotation} & \textbf{Among} & \textbf{Around} & \textbf{Overall ↑} \\ \hline
    w/o $\mathcal{L}_{3D}$                                            & 35.0              & 55.1           & 66.8            &  54.1              \\
    \rowcolor[HTML]{E7E6E6}
    Full                                      & \textbf{44.0}              & \textbf{64.8}  & \textbf{72.4}   & \textbf{62.7}     \\ \hline
    \end{tabular}
\end{table}
To evaluate the impact of $\mathcal{L}_{3D}$ on our framework, we further conduct the ablation where the 3D foundation model was removed, and only the cross-entropy loss ($\mathcal{L}_{\text{text}}$) was applied to tokens other than the 3D special token. As shown in Tab.~\ref{table:3dloss}, using Qwen2.5-VL-3B as an example, the performance drops from 62.7 (full) to 54.1 without $\mathcal{L}_{3D}$, though it remains slightly higher than the CoT SFT (53.4). These results highlight two key insights: (1) \emph{incorporating the 3D foundation model helps the VLM utilize the special token to encode geometric information}, thereby enhancing the spatial reasoning capability; and (2) even without the 3D foundation model, \emph{introducing special tokens increases the representational flexibility of the model}, leading to moderate improvements on certain metrics.

\section{Training Curve}\label{sec:curve}
\begin{figure}[]
    \centering
    \includegraphics[width=0.48\textwidth]{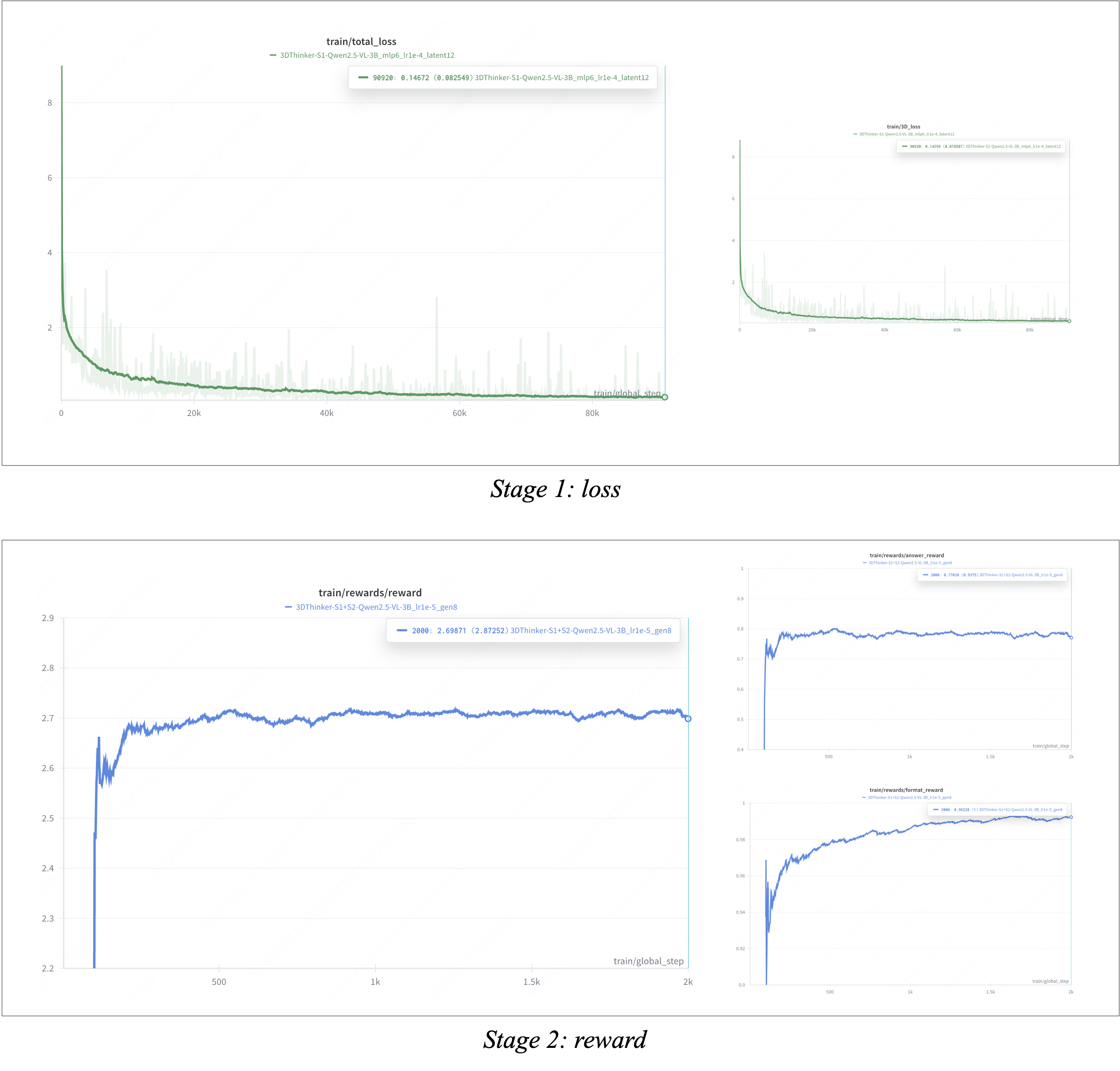}
    \vspace{-0.8em}
    \caption{Visualization of our training curves in terms of Qwen2.5-VL-3B on stage 1 (\textcolor{green}{green} curves) and stage 2 (\textcolor{blue}{blue} curves). Top-left: total loss of stage 1; Top-right: 3D loss of stage 1; Bottom-left: total reward of stage 2; Bottom-right: answer reward and format reward of stage 2.}\label{fig:curve}
\end{figure}
As shown in Fig.~\ref{fig:curve}, we visualize the training curves of our two-stage process in terms of training on MindCube-only. The loss in stage 1 (\textcolor{green}{green}) converges after approximately 20k steps, reaching around 0.15. The reward in stage 2 (\textcolor{blue}{blue}) converges much faster, stabilizing after roughly 500 steps with a reward value of about 2.7. In the bottom-right panel, we further plot the curves of the format reward ($r_{\text{format}}$) and the answer reward ($r_{\text{ans}}$). Both exhibit a brief decline during the early phase of training, followed by a steady upward trend. \emph{This initial drop can be attributed to the model’s exploration behavior at the beginning of reinforcement learning}, during which it searches the solution space and may temporarily follow suboptimal trajectories before discovering more optimal ones that lead to rapid performance improvement.

\section{Details for Dataset Generation}\label{sec:instruction}
\begin{figure}[!htbp]
    \centering
    \includegraphics[width=0.45\textwidth]{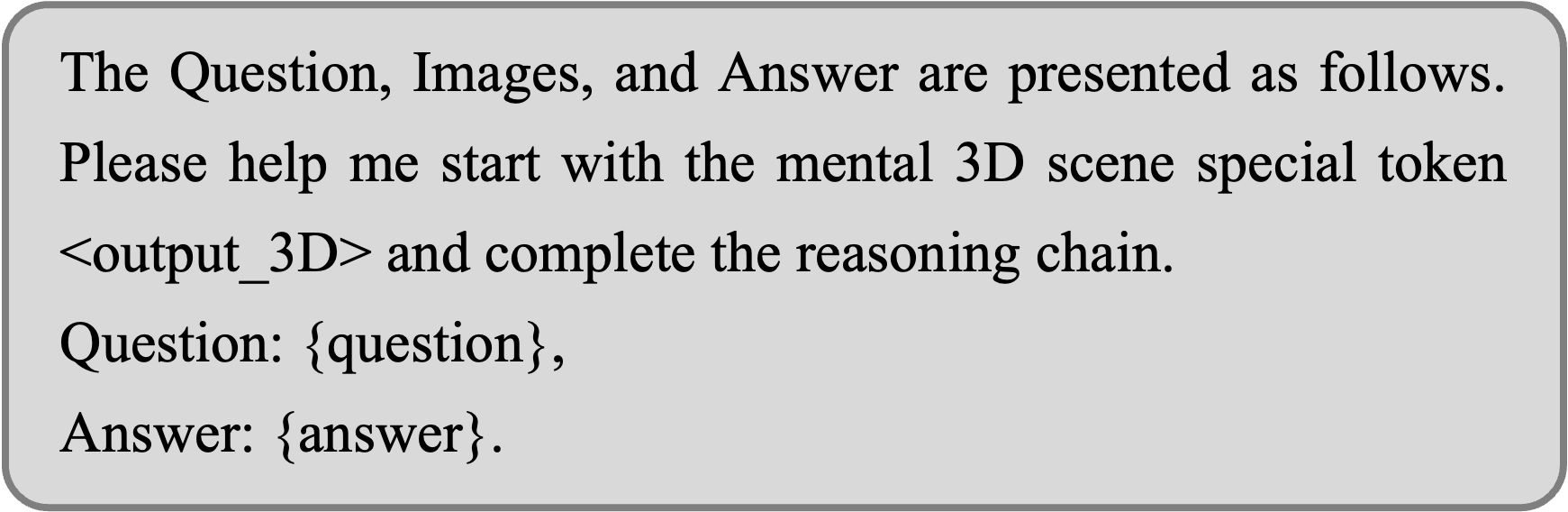}
    \caption{Prompt for the CoT generation, with the input of question and GT answer.}\label{fig:prompt}
\end{figure}
\begin{figure}[!htbp]
    \centering
    \includegraphics[width=0.47\textwidth]{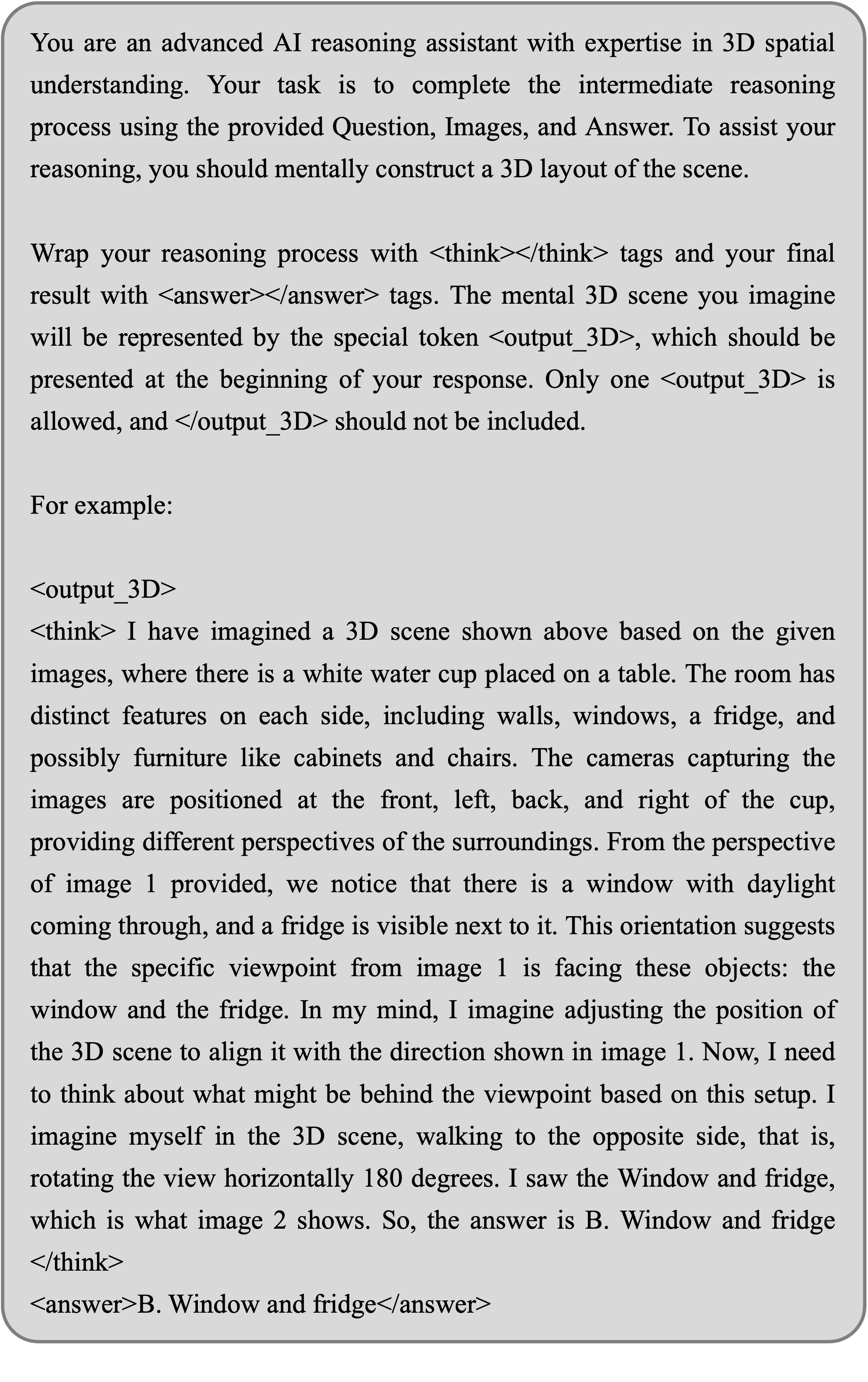}
    \caption{System prompt for the CoT generation .}\label{fig:system_prompt}
\end{figure}
For the training dataset, which consists of image set from different views $I$, question $Q$ and GT response $R$, we employed GPT-4.1 ($M$) to complete the CoT reasoning process $o$. Fig.~\ref{fig:prompt} and Fig.~\ref{fig:system_prompt} illustrate the prompts we designed for generating CoT data. Additionally, we assess the quality of the generated CoT, ensuring that $o$ contains ${\small \texttt{<output\_3D>}}$ at the beginning, followed by the text in adherence to the format ${\small \texttt{<think>...</think> <answer>...</answer>}}$. The CoT data $o$ that do not conform to this specification will be re-generated by $M$. If the data fails to meet the required format after more than $N=10$ attempts, it will be discarded.

\section{Explanation of the Claim}\label{sec:claim}
\noindent\textbf{Annotation-free}: In fact, unlike previous purely textual CoT methods, our method requires 3D mentaling during the reasoning process. Thus, “annotation-free” refers specifically to the absence of \emph{manual geometric annotations for 3D mentaling} (e.g., ground-truth point clouds). In our framework, GPT-4.1 is only used to generate 3D placeholders, while the actual 3D geometric features associated with these placeholders are provided by VGGT aggregator.

\noindent\textbf{VGGT Dependence}:  In the supervised training stage, we aim to elicit ``think with 3D'' behaviour without relying on manual geometric annotations, which leads us to make use of features extracted by VGGT. In contrast, learning to generate 3D geometry entirely from scratch would necessitate explicit geometric supervision. On the other hand, in the reinforced training stage, learning can be driven \emph{solely by the outcome-based reward signal}. Specifically, as shown in Tab. 6 of the main text, even in the absence of $r_{\text{3D}}$ there is still an \textbf{8.93\%} (68.3 vs. 62.7) improvement, indicating that \method \emph{remains robust without the teacher model}.

\section{Cost}\label{sec:cost}
Our cost reduction primarily stems from two factors: 
(1) \method typically reaches near-optimal performance within 500 training steps (see Fig.~\ref{fig:curve}); (2) with VLLM acceleration, RL rollout efficiency is substantially improved. Consequently, taking Qwen2.5-VL-3B as an example, on a single H200 GPU with a batch size of 1, the convergence time for supervised training and reinforcement training is \textbf{21.84 h} and \textbf{12.85 h}, respectively.

\section{More Visualization}\label{sec:visualization}
\subsection{Thinking with 3D Visualization}\label{sec:failure}
\begin{figure*}[]
    \centering
    \includegraphics[width=\textwidth]{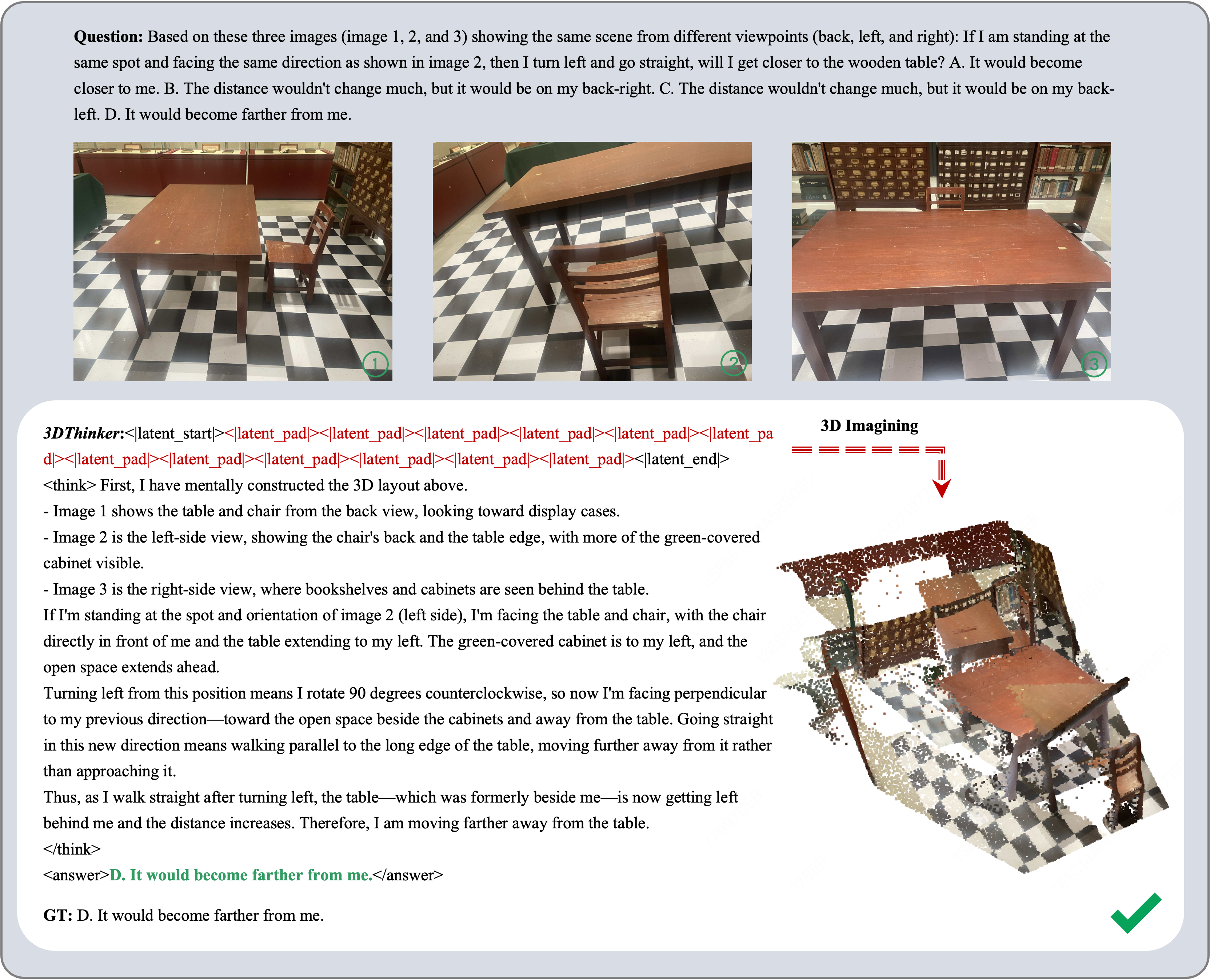}
    \vspace{-0.8em}
    \caption{More visualization results of our method. The reasoning process, GT answer, and the imagery of reconstructed point cloud from 3D latent are presented.}\label{fig:vis1}
\end{figure*}

\begin{figure*}[]
    \centering
    \includegraphics[width=\textwidth]{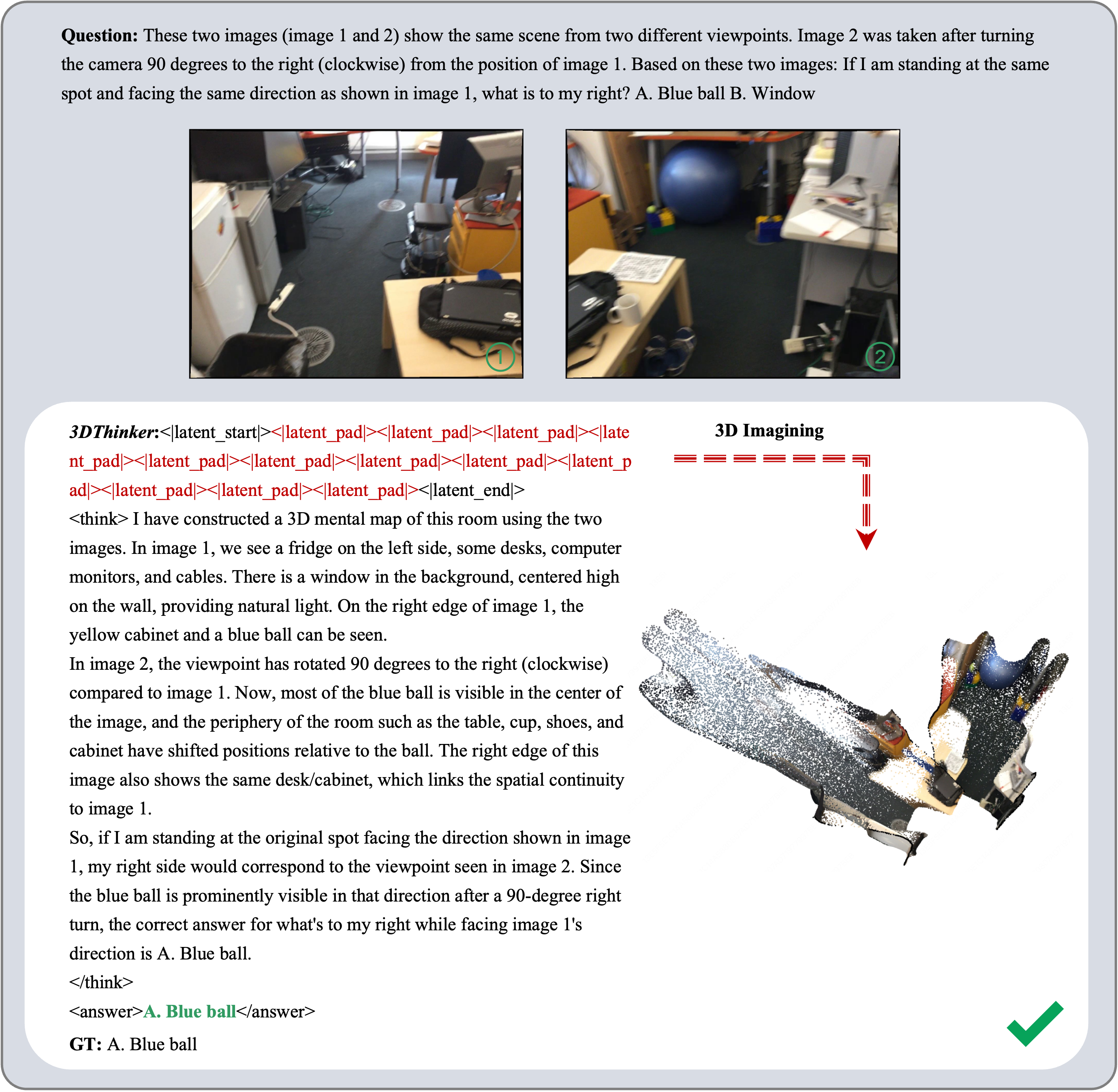}
    \vspace{-0.8em}
    \caption{More visualization results of our method. The reasoning process, GT answer, and the imagery of reconstructed point cloud from 3D latent are presented.}\label{fig:vis2}
\end{figure*}

\begin{figure*}[]
    \centering
    \includegraphics[width=\textwidth]{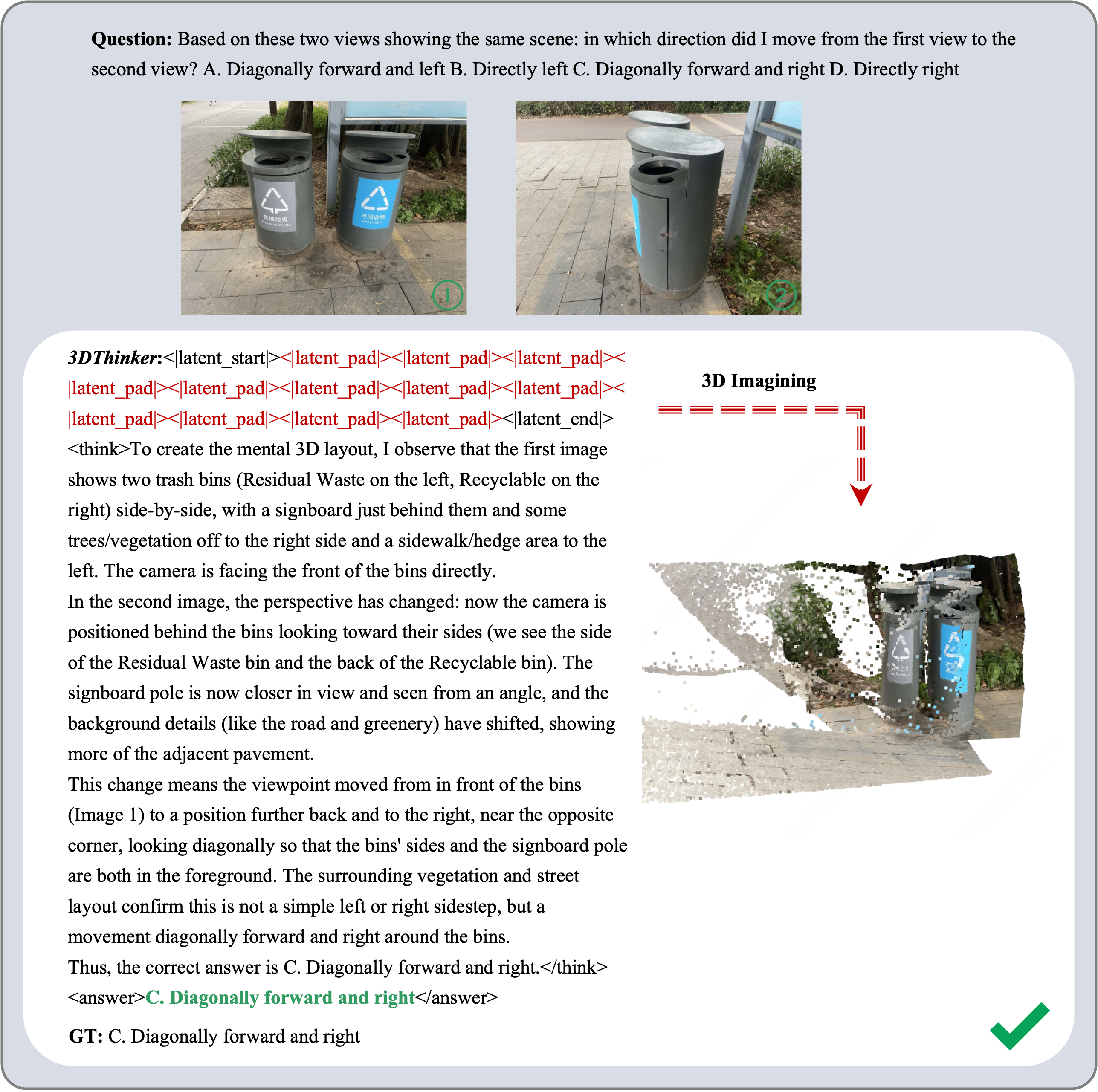}
    \vspace{-0.8em}
    \caption{More visualization results of our method. The reasoning process, GT answer, and the imagery of reconstructed point cloud from 3D latent are presented.}\label{fig:vis3}
\end{figure*}

\begin{figure*}[]
    \centering
    \includegraphics[width=\textwidth]{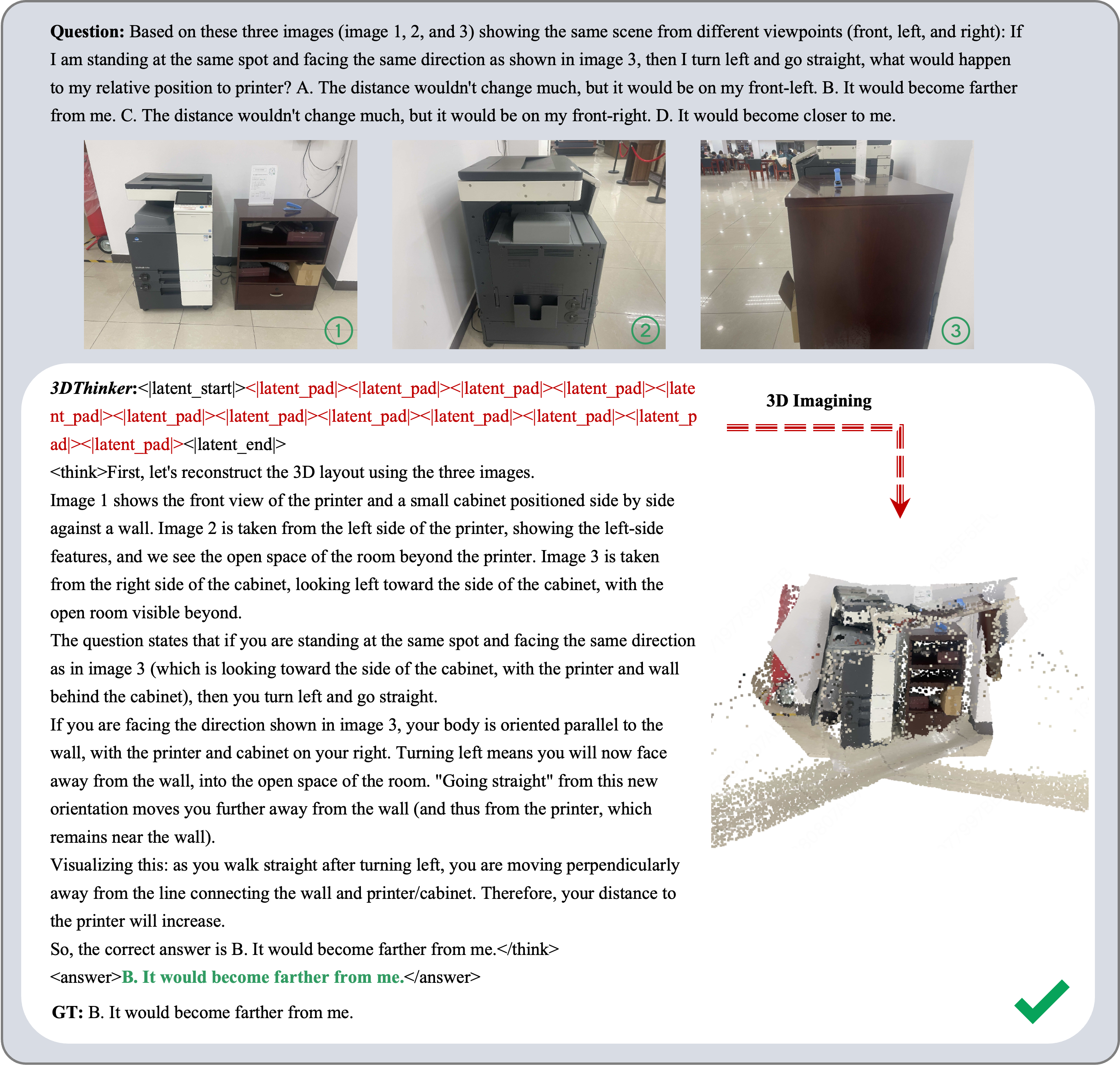}
    \vspace{-0.8em}
    \caption{More visualization results of our method. The reasoning process, GT answer, and the imagery of reconstructed point cloud from 3D latent are presented.}\label{fig:vis4}
\end{figure*}

\begin{figure*}[]
    \centering
    \includegraphics[width=\textwidth]{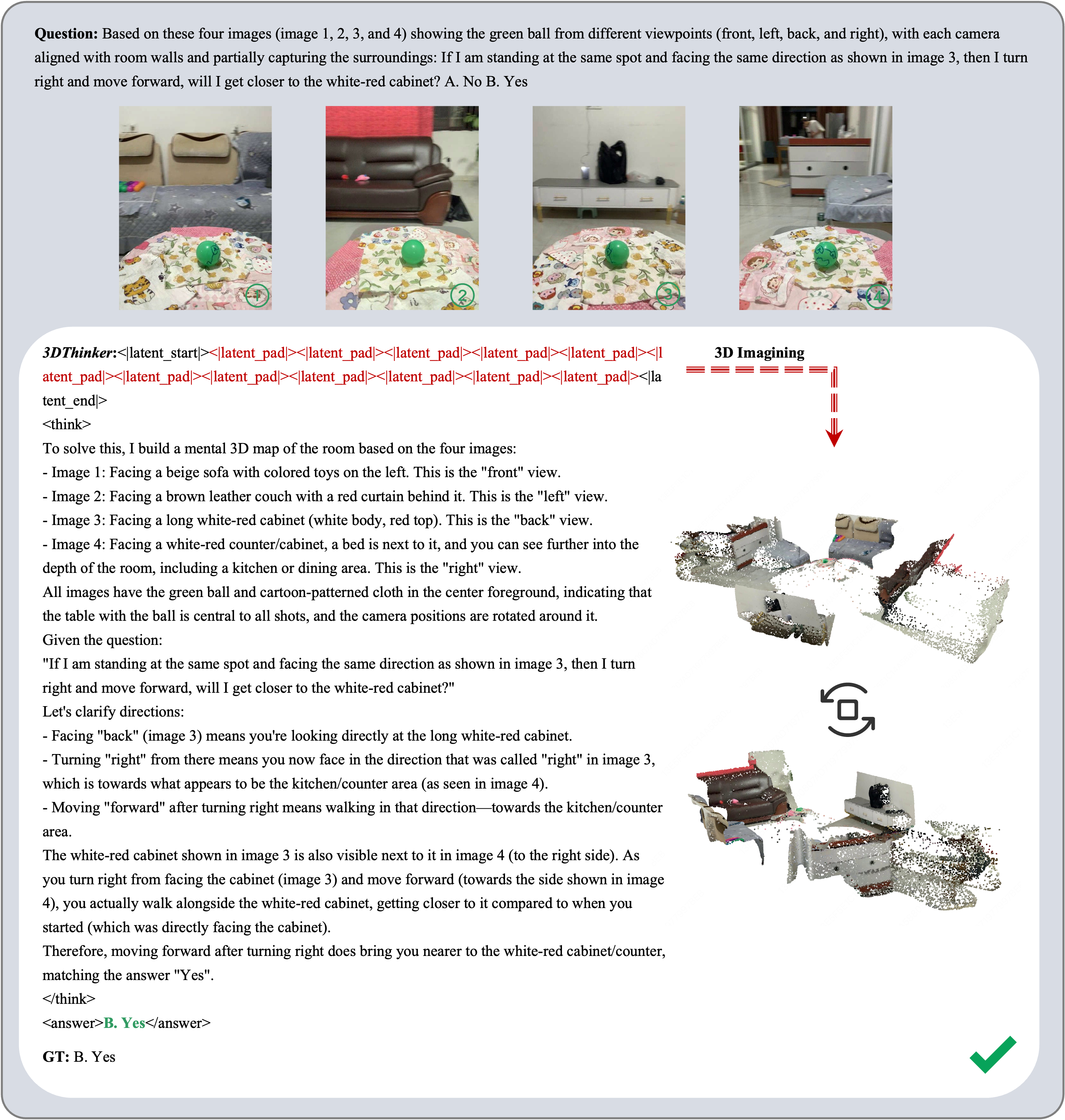}
    \vspace{-0.8em}
    \caption{More visualization results of our method. The reasoning process, GT answer, and the imagery of reconstructed point cloud from 3D latent are presented.}\label{fig:vis5}
\end{figure*}

\begin{figure*}[]
    \centering
    \includegraphics[width=\textwidth]{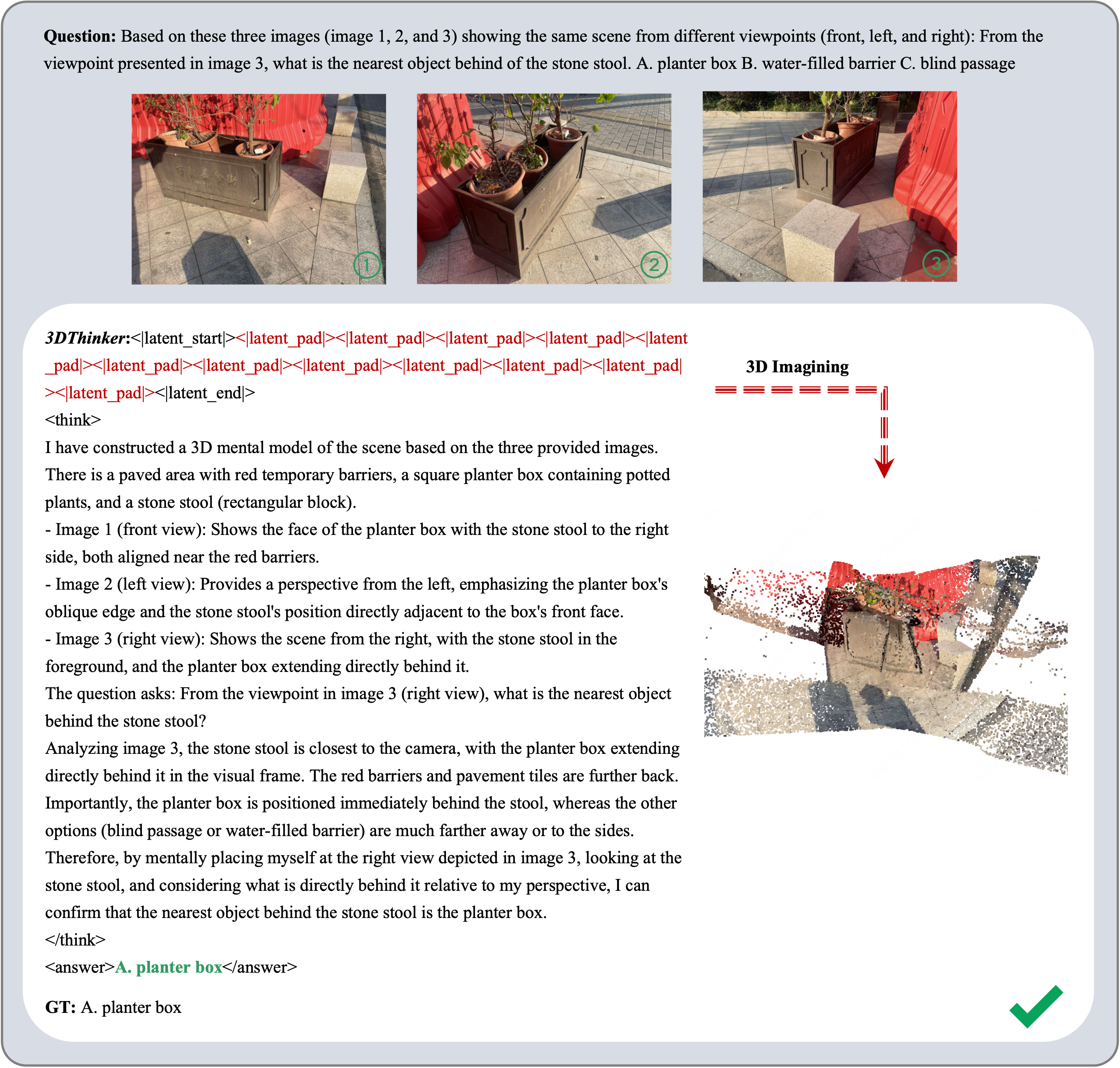}
    \vspace{-0.8em}
    \caption{More visualization results of our method. The reasoning process, GT answer, and the imagery of reconstructed point cloud from 3D latent are presented.}\label{fig:vis6}
\end{figure*}

\begin{figure*}[]
    \centering
    \includegraphics[width=\textwidth]{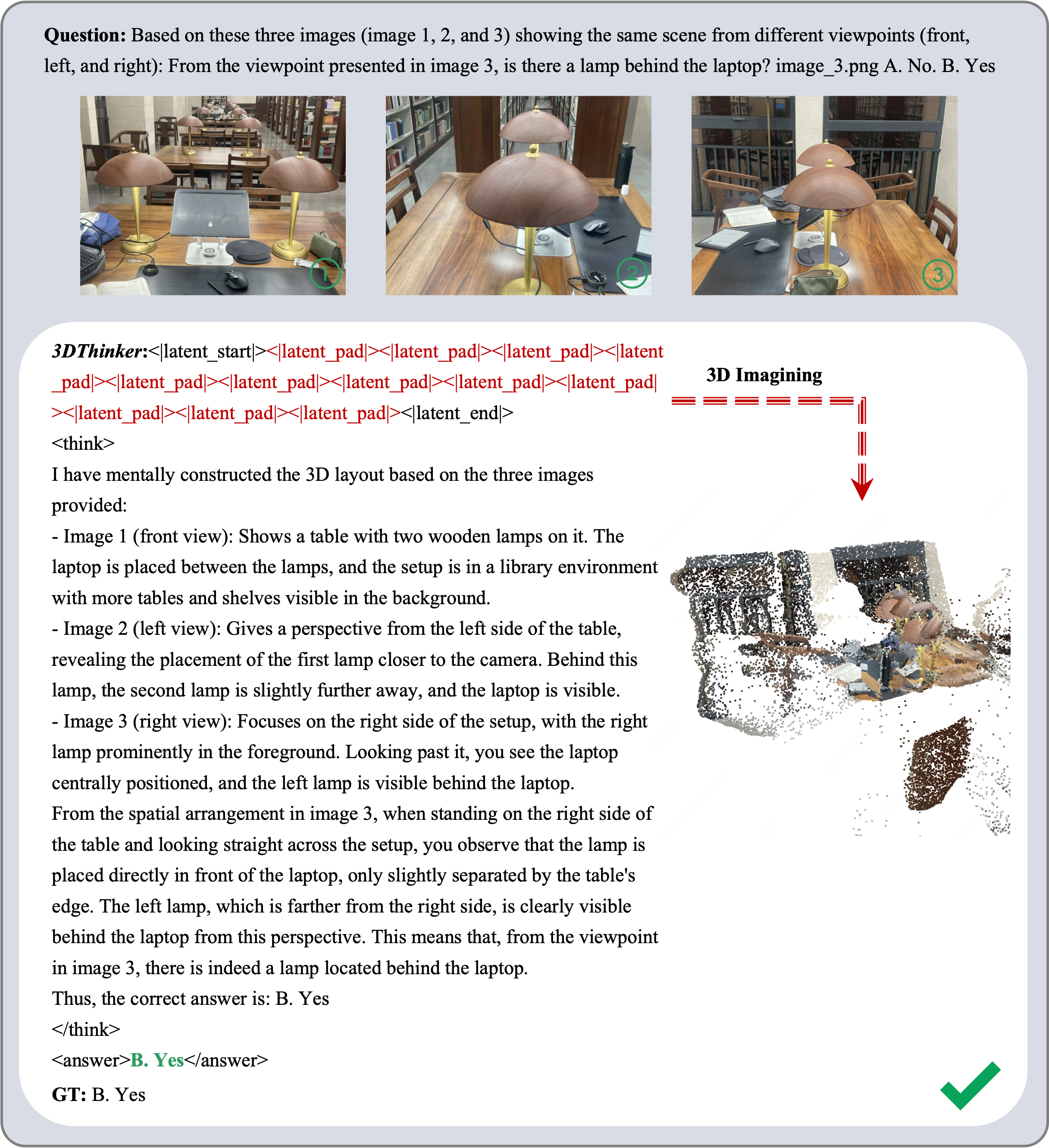}
    \vspace{-0.8em}
    \caption{More visualization results of our method. The reasoning process, GT answer, and the imagery of reconstructed point cloud from 3D latent are presented.}\label{fig:vis7}
\end{figure*}

In Fig.~\ref{fig:vis1}, Fig.~\ref{fig:vis2}, Fig.~\ref{fig:vis3}, Fig.~\ref{fig:vis4}, Fig.~\ref{fig:vis5}, Fig.~\ref{fig:vis6}, and Fig.~\ref{fig:vis7}, we provide additional visualizations on MindCube-Tiny of $\textit{\method-S1+S2}_\textnormal{Qwen2.5-3B}$, including the reasoning output, 3D imagery, etc.. \emph{The 3D imagery (reconstructed 3D point cloud) shows what the VLM is “thinking” during spatial reasoning}.
\subsection{Failure Case and Limitation}\label{sec:failure}
\begin{figure*}[]
    \centering
    \includegraphics[width=\textwidth]{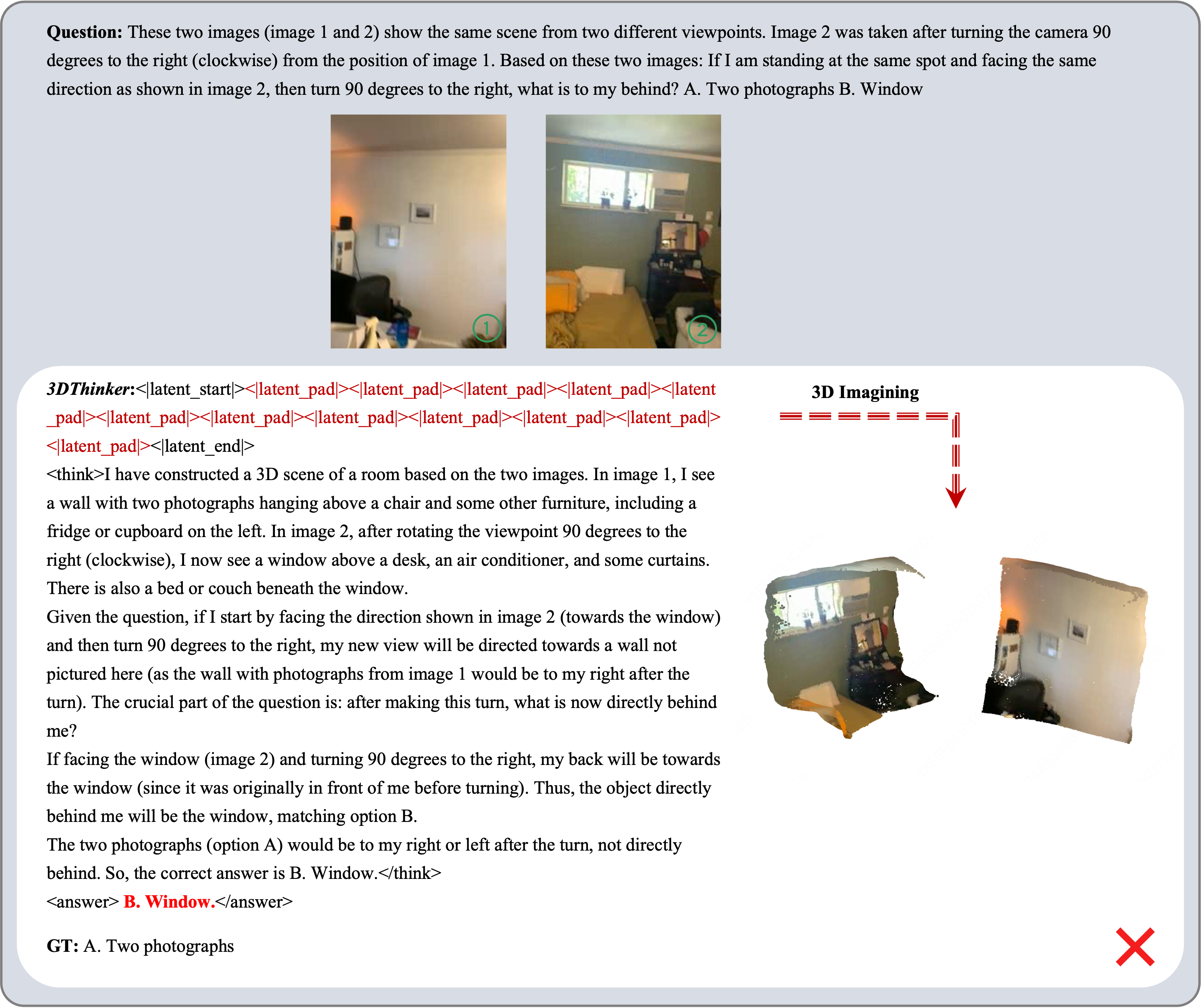}
    \vspace{-0.8em}
    \caption{Visualization of the failure case. The reasoning process, GT answer, and the imagery of reconstructed point cloud from 3D latent are presented.}\label{fig:vis_fail}
\end{figure*}

As shown in Fig.~\ref{fig:vis_fail}, we present a representative failure case. Given two images from the indoor scene, \method incorrectly infers the spatial position of the rear object. Fortunately, by visualizing the 3D mental representation, we can identify the underlying cause. The reconstructed 3D point cloud exhibits structural inconsistencies—specifically, \emph{the wall with two photographs is mistakenly placed on the right side of the frontal window view, whereas it should appear on the left}. This misalignment likely stems from the sparsity of the two input views, which \emph{provide insufficient spatial constraints, leading to inaccurate 3D reconstruction and, consequently, erroneous spatial reasoning}. Therefore, in the case of extremely sparse views, better integration of other information is required, which could serve as a promising direction for future work.

\end{document}